%% file: main.tex
\documentclass[twoside,11pt]{article}

\usepackage{jmlr2e}
\usepackage{graphicx}
\usepackage{amsmath}
\usepackage{amssymb}
\usepackage{float}

\usepackage{caption}
\usepackage{subcaption}
\usepackage{dirtytalk}

\newcommand{\norm}[1]{\left\lVert#1\right\rVert}

\DeclareMathOperator{\rank}{rank}

\arxivheading

\ShortHeadings{Review of Structured CNN compression} 
{Kuzmin, Nagel, Pitre, Pendyam, Blankevoort and Welling}

\firstpageno{1}

\begin{document}

\title{Taxonomy and Evaluation of Structured Compression of Convolutional Neural Networks}

\author{\name Andrey Kuzmin \email akuzmin@qti.qualcomm.com \\
       \addr Qualcomm AI Research\thanks{Qualcomm AI Research is an initiative of Qualcomm Technologies, Inc.}\\
       Qualcomm Technologies Netherlands B.V.
       \AND
       \name Markus Nagel \email markusn@qti.qualcomm.com \\
       \addr Qualcomm AI Research\footnotemark[1]\\
       Qualcomm Technologies Netherlands B.V.
       \AND
       \name Saurabh Pitre \email spitre@qti.qualcomm.com \\
       \addr Qualcomm AI Research\footnotemark[1]\\
       Qualcomm Technologies, Inc.
       \AND
       \name Sandeep Pendyam \email spendyam@qti.qualcomm.com \\
       \addr Qualcomm AI Research\footnotemark[1]\\
       Qualcomm Technologies, Inc.
       \AND
       \name Tijmen Blankevoort \email tijmen@qti.qualcomm.com \\
       \addr Qualcomm AI Research\footnotemark[1]\\
       Qualcomm Technologies Netherlands B.V.
       \AND
       \name Max Welling \email mwelling@qti.qualcomm.com \\
       \addr Qualcomm AI Research\footnotemark[1]\\
       Qualcomm Technologies Netherlands B.V.}
\editor{}

%

\maketitle

\begin{abstract}
The success of deep neural networks in many real-world applications is leading to new challenges in building more efficient architectures. One effective way of making networks more efficient is neural network compression. We provide an overview of existing neural network compression methods that can be used to make neural networks more efficient by changing the architecture of the network. First, we introduce a new way to categorize all published compression methods, based on the amount of data and compute needed to make the methods work in practice. These are three `levels of compression solutions'. Second, we provide a taxonomy of tensor factorization based and probabilistic compression methods. Finally, we perform an extensive evaluation of different compression techniques from the literature for models trained on ImageNet. We show that SVD and probabilistic compression or pruning methods are complementary and give the best results of all the considered methods. We also provide practical ways to combine them. 
\end{abstract}


\section{Introduction}
\input{sections/introduction.tex}

\section{Related work}
\input{sections/related_work.tex}

\input{sections/methodology.tex}

\section{Structured compression methods overview}
\input{sections/methods_level_1.tex}


\input{sections/methods_level_2.tex}
\input{sections/methods_level_3.tex}

\input{sections/compression_ratio_selection.tex}

\input{sections/experiments.tex}

\section{Conclusion}
\input{sections/conclusion.tex}

\section*{Acknowledgments}
We would like to thank Arash Behboodi, Christos Louizos and Roberto Bondesan for their helpful discussions and valuable feedback.

\vskip 0.2in
\bibliography{references}

\end{document}

%% file: sections/introduction.tex
Due to the tremendous success of deep learning, neural networks can now be found in applications everywhere. Running in the cloud, on-device, or even on dedicated chips, large deep learning networks now form the foundation for many real-world applications. They are found in voice assistants, medical image analyzers, automatic translation tools, software that enhances photographs, and many other applications. 

In these real-world applications, the performance of neural networks is an important topic. Well-performing deep neural networks are large and expensive to execute, restricting their use in, e.g., mobile applications with limited compute. Even for large-scale cloud-based solutions, such as services that process millions of images or translations, neural network efficiency directly impacts compute and power costs.

Alongside quantization (\cite{krishnamoorthi}) and optimizing kernels for efficient deep learning execution (\cite{cudnn}), neural network compression is an effective way to make the run-time of these models more efficient. With compression, we mean improving the run-time of models, as opposed to compressing the actual size of the network for storage purposes. In this paper, we will describe and compare several methods for compressing large deep-learning architectures for improved run-time.

Even for architectures that were designed to be efficient, such as MobilenetV2 (\cite{sandler2018mobilenetv2} and EfficientNet (\cite{efficientnet}), it is still helpful to do neural network compression \citep{liu2019metapruning, he2017channel}. There has been a debate in the deep-learning literature on the efficacy of compression. \citet{rethinking} argues that network compression does not help, and one could have trained that similar architecture from scratch. However, the ``The lottery-ticket hypothesis", \citet{lottery} provides arguments for the hypothesis that it's better to train a large network and compress it, rather than training a smaller model from scratch. We will see in our result section more evidence for the latter, indicating it helps to compress networks after training, as opposed to starting with a more efficient architecture.

In this paper, we systematically categorize the many different compression methods that have been published and test all of them on a large scale image classification task. We group methods by their practical usage into 3 different levels. Level 1: Methods that do not use data. Level 2: methods that do not use back-propagation and Level 3: methods that use a training procedure. Within these categories, we look at several different ways of doing neural network compressing, including tensor-decomposition, channel-pruning, and several Bayesian inspired approaches. Specifically, we look only at structured pruning approaches, where the size of the tensors of the network decreases in size. This is opposed to unstructured pruning methods, such as (\cite{han2015deep}) and (\cite{molchanov2017variational}), that remove individual weights from the network. These types of pruning methods require specific hardware to obtain speed-ups, whereas structured pruning methods more directly provide improved speed on most devices.

%% file: sections/related_work.tex
\paragraph{SVD-based methods.}
SVD decomposition was first used by~\cite{denil2013predicting} to demonstrate redundancy in weight parameters in deep neural networks. Following this approach, several works employ low-rank filter approximation~\citep{jaderberg2014speeding, denton2014exploiting} to reduce inference time for pre-trained CNN models. One of the first methods for accelerating convolutional layers by applying low-rank approximation to the kernel tensors is~\cite{denton2014exploiting}. The authors suggest several decompositions approaches applied to parts of the kernel tensor obtained by bi-clustering. The spatial decomposition method from \cite{jaderberg2014speeding} decomposes a $k\times k$ filter into a $k \times 1$  and $1\times k$ while exploiting redundancy among multiple channels.
Another notable improvement of SVD compression is reducing the error introduced by filter approximation based on input data. The approach suggested by~\cite{zhang2016accelerating} uses per-layer minimization of errors in activations for compressed layers. 

\paragraph{Tensor decomposition-based methods.}
Several approaches for structured CNN compression based on tensor decomposition applied to 4D convolutional kernels were suggested. An overview of tensor decomposition techniques is given in~\cite{kolda2009tensor}. The authors of~\cite{lebedev2014speeding} apply CP-decomposition to compress a kernel of a convolutional filter. The work of~\cite{kim2015compression} suggests a CNN compression approach based on the Tucker decomposition. The authors also suggest employing analytic solutions for variational Bayesian matrix factorization (VBMF) by~\cite{nakajima2013global} for the rank selection. Another tensor decomposition approach that was applied to model compression is the tensor-train decomposition~\citep{oseledets2011tensor}. It is used in~\cite{novikov2015tensorizing} for compression of fully-connected layers, and~\cite{garipov2016ultimate} applies it for convolutional layers. 

Another direction in convolutional layer compression is to increase the dimensionality by reshaping a kernel into a higher-dimensional tensor~\citep{su2018tensorized, novikov2015tensorizing}. For example, a $3 \times 3$ convolutional kernel with 64 input, and 64 output channels represented as 6-dimensional $8 \times 8 \times 8 \times 8 \times 3 \times 3$ tensor instead $64 \times 64 \times 3 \times 3$, where $8 \times 8$ corresponds to one way of factorizing 64. Using any other way of factorizing 64 in combination with any of the three tensor decomposition techniques yields a new compression technique. An extensive study of applying CP-decomposition, Tucker decomposition, and tensor-train decomposition in combination with factorizing kernel dimensions were published by~\cite{su2018tensorized}. They consider compression of both fully-connected and convolutional layers.

\paragraph{Pruning methods.}
One of the ways to reduce inference time for pre-trained models is to prune redundant channels. The work of~\cite{li2016pruning} is focused on using channel norm magnitude as a criterion for pruning. Another approach is to use a lasso feature selection framework for choosing redundant channels while minimizing reconstruction error for the output activation based on input data~\citep{he2017channel}.





\paragraph{Compression ratio selection methods.}
As every layer of a neural network has different sensitivity to compression, any SVD or tensor decomposition technique can be further improved by optimizing per layer compression ratios. The methods~\cite{kim2019efficient,kim2018automatic} suggest efficient search strategies for the corresponding discrete optimization problem. A learning-based strategy based on reinforcement learning is suggested in~\cite{he2018amc}. 

\paragraph{Loss-aware compression.}
While compression and pruning methods reduce complexity, most of the methods assume equal importance of every model parameter for the accuracy of the final model. One way to improve compression methods is to estimate the importance of each of the weights, and use this information while pruning. Several methods suggest introducing importance based on loss function increase~\citep{wang2019eigendamage, gao2018rate, lecun1990optimal, hassibi1993optimal}. 
The increase of the loss function is often estimated based on first or second-order linear approximations. 

\paragraph{Probabilistic compression.}
Another family of methods from the literature suggests adding a term to a loss function that controls the complexity of the model. Typically, a collection of stochastic gates is included in a network, which determines which weights are to be set to zero. Methods following this approach include~\cite{christosl0,neklyudov,vibnet}, and a recent survey is provided in~\cite{gale2019state}.

\paragraph{Efficient architecture design.} Several works aim at finding the optimal trade-off between model efficiency and prediction accuracy. MobileNet V1~\citep{howard2017mobilenets} is based on combining depth-wise separable convolutions and depth-wise convolutions to reduce the number of FLOPs. MobileNet V2~\citep{sandler2018mobilenetv2} is based on the linear bottleneck and inverted residual structure and further improves the efficiency of the model. MnasNet~\citep{tan2019mnasnet} is based on a combination of squeeze and excitation blocks. Another efficient architecture~\citep{zhang2018shufflenet} leverages group convolution, and channel shuffle operations. Some of the more recent architectures~\citep{mobilenetV3,wu2019fbnet} are based on combining efficient handcrafted layers with neural architecture search.

%% file: sections/methodology.tex
\subsection{Levels of compression solutions}

To facilitate a comparison of the methods proposed in the literature, we refer to practical use cases of model compression. The following levels of compression solutions are introduced in a way similar to~\cite{nagel2019data}. The definition of each level depends on the amount of training data and computational resources available when using a compression method. 

\begin{figure}[t]
\centering
\begin{small}
\begin{tabular}{c}
\includegraphics[trim={0.0 2.5cm 0 0},clip, width=9.0cm]{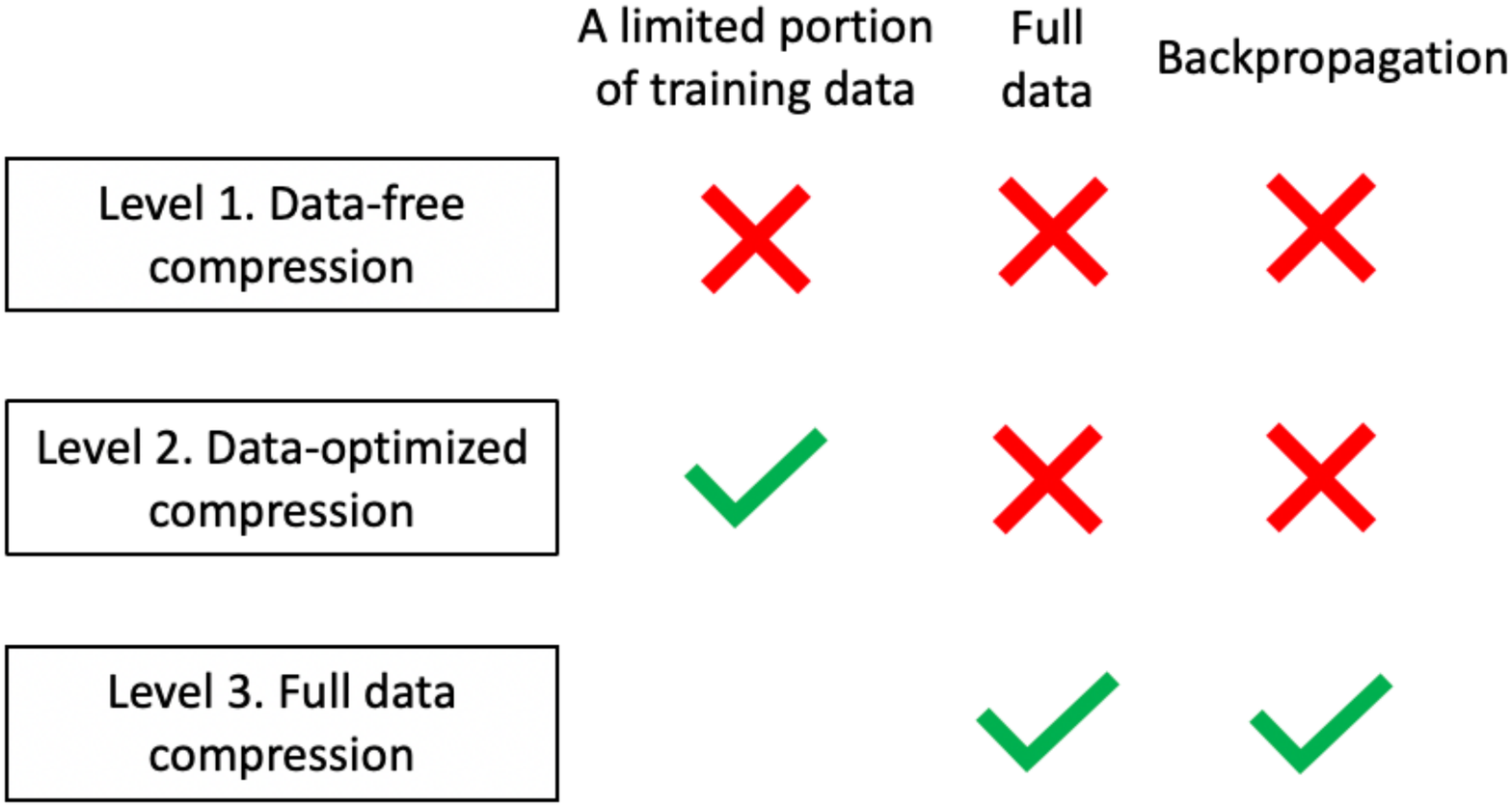}
\label{fig:methodology_levels}
\end{tabular}
\end{small}
\caption{Levels used for comparison of the model compression methods.}

\end{figure}

\begin{itemize}
    \item \textit{Level 1. Data-free compression}. No data or training pipeline is available in this case. Nevertheless, the goal is to produce an efficient model with the predictions as close to the original model as possible.
    \item \textit{Level 2. Data-optimized compression}. A limited number of batches of the training data are used to guide the compression method with no ground-truth labels being used. In this case, layer-wise optimization of the parameters of the compressed model is used to improve the predictions. No back-propagation is used at this level.
    \item \textit{Level 3. Full data compression}. This level corresponds to fine-tuning of the compressed model using the full training set or training an efficient model from scratch using the full amount of data. Full back-propagation is used in this case so that the computational complexity is comparable to the complexity of the original model training procedure.
\end{itemize}
Different from~\cite{nagel2019data}, in the current work we omit introducing one more level for the methods which introduce architecture changes, as compression is complementary to architecture search methods and allows to obtain further performance improvement even if applied for handcrafted or learning based efficient architectures~\cite{he2018amc,liu2019metapruning}.

The compression levels are summarized in figure~1. Using the levels formulation, all the compression methods can be categorized and compared in a similar setting. The practical choice of compression level depends on the specific envisioned use case.

%% file: sections/methods_level_1.tex
To define a quantitative measure of compression, we use the number of multiply-accumulate operations (MAC units, or MACs) used by a neural network at inference time. Given a network with $L$ layers with $c_i$ operations in each, the total computational complexity $C$ is expressed as:
\begin{equation}
C = \sum_{i=1}^{L} c_i.    
\end{equation}
Assuming that a compression technique reduces the number of operations per layer to $\hat{c}_i$, per layer compression ratio $\alpha_i$ can be computed as:
\begin{equation}
\alpha_i = 1 - \frac{\widehat{c_i}}{c_i}.    
\end{equation}
The whole model compression rate $\alpha$ can be defined in a similar way:
\begin{equation}
\alpha = 1 - \frac{\widehat{C}}{C},   
\end{equation}
where $\widehat{C}$ is the total number of operations in the compressed model. 

In practice, the model's accuracy has a different sensitivity to the compression of different layers. The problem of selecting an optimal compression ratio for each of the layers given the target whole-model compression ratio is considered in section \ref{sec:compression_ratio_selection}. 

\subsection{Level 1. Data-free compression methods}
A convolutional layer is specified by the kernel $\mathbf{W} \in \mathbb{R}^{k \times k \times s \times t}$, where $s$ is the number of input channels, $t$ is the number of output channels and the spatial size of the filter is $k \times k$. The kernel is assumed to be square of odd size $k$ for simplicity, $\delta=(k-1)/2$ denotes its \say{half-width}.
A convolution is a linear transformation of the feature map $\mathbf{X} \in \mathbb{R}^{s \times w \times h}$ into an output tensor $\mathbf{Y} \in \mathbb{R}^{t \times w \times h}$. We assume the spatial dimensions of the input and output feature maps are equal in order to avoid notational clutter. The convolution is defined as follows:
\begin{equation}
\begin{split}
Y(i_t,i_x,i_y) = 
\sum_{i_s=1}^{s} \sum_{i_x^\prime=i_x-\delta}^{i_x+\delta} \sum_{i_y^\prime=i_y-\delta}^{i_y+\delta} W(i_x^\prime-i_x+\delta,i_y^\prime-i_y+\delta,i_s,i_t) X(i_s,i_x^\prime,i_y^\prime).
\label{eq:conv_orig}
\end{split}
\end{equation} 
We omit the bias term for notation simplicity. The number of MACs in a convolutional layer is $c=k^2sthw$.

\subsubsection{SVD methods}
\label{section:svd}
To leverage low-rank matrix approximation for the compression of a convolutional layer, the kernel tensor is transformed into a matrix. In this case, the dimensions of a tensor are referred to as \textit{modes}. There are seven types of possible matricizations of a 4-dimensional tensor. Two of these are used in the compression methods that are introduced in the following paragraphs. 

\paragraph{Weight SVD.} This method is based on reshaping the kernel tensor into a matrix $\mathbf{W} \in \mathbb{R}^{k^2s \times t}$ followed by a low-rank approximation. This type of matricization corresponds to merging three of the four original modes $k \times k \times s \times t$ into a single supermode.

\begin{figure*}
\label{fig:conv_svd_diags}
\centering
\begin{small}
\begin{tabular}{ccc}
\vtop{%
\vskip6pt
  \hbox{%
    \includegraphics[height=2.0cm]{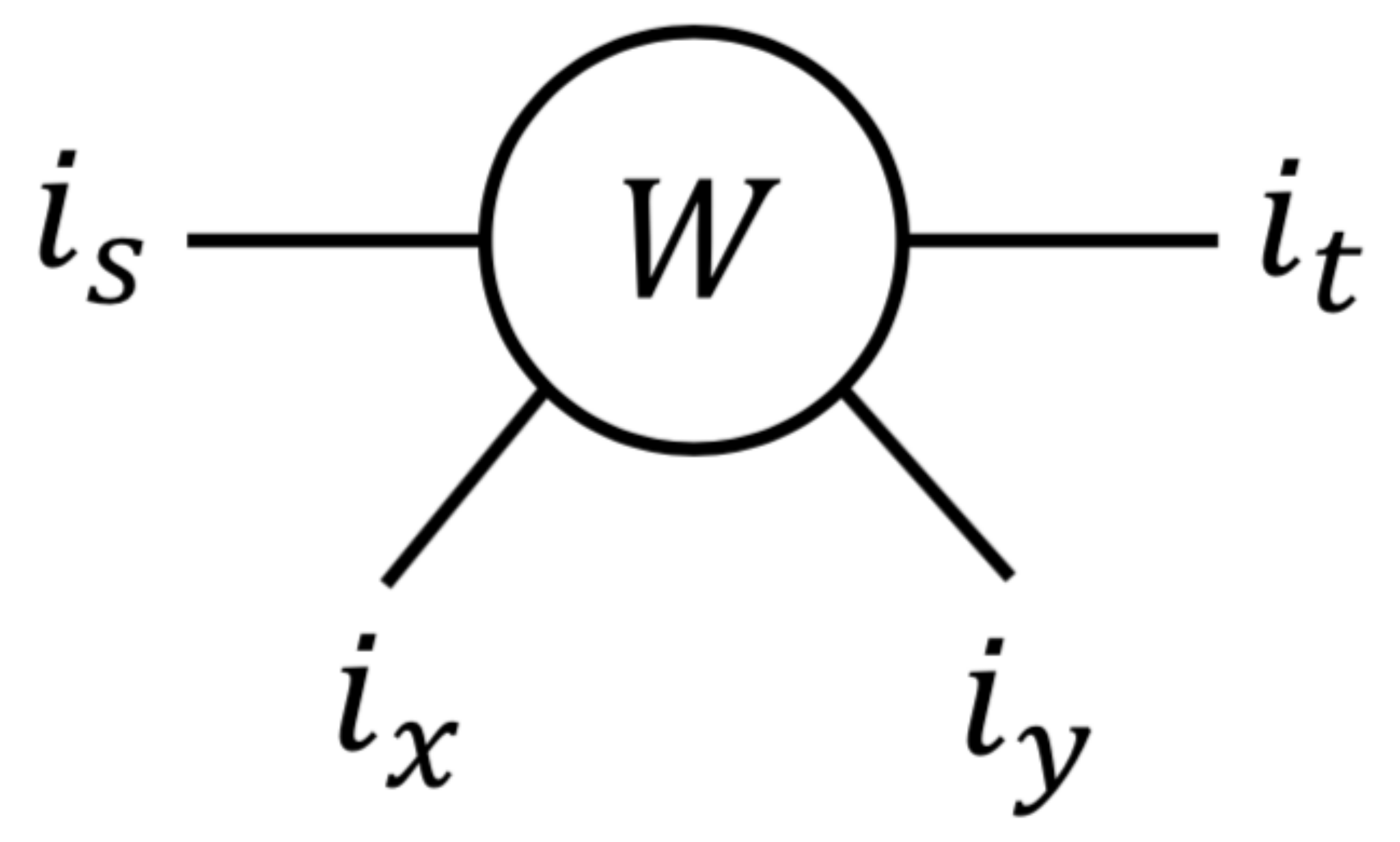}%
  }%
}
&
\vtop{%
\vskip1pt
  \hbox{%
    \includegraphics[height=2.1cm]{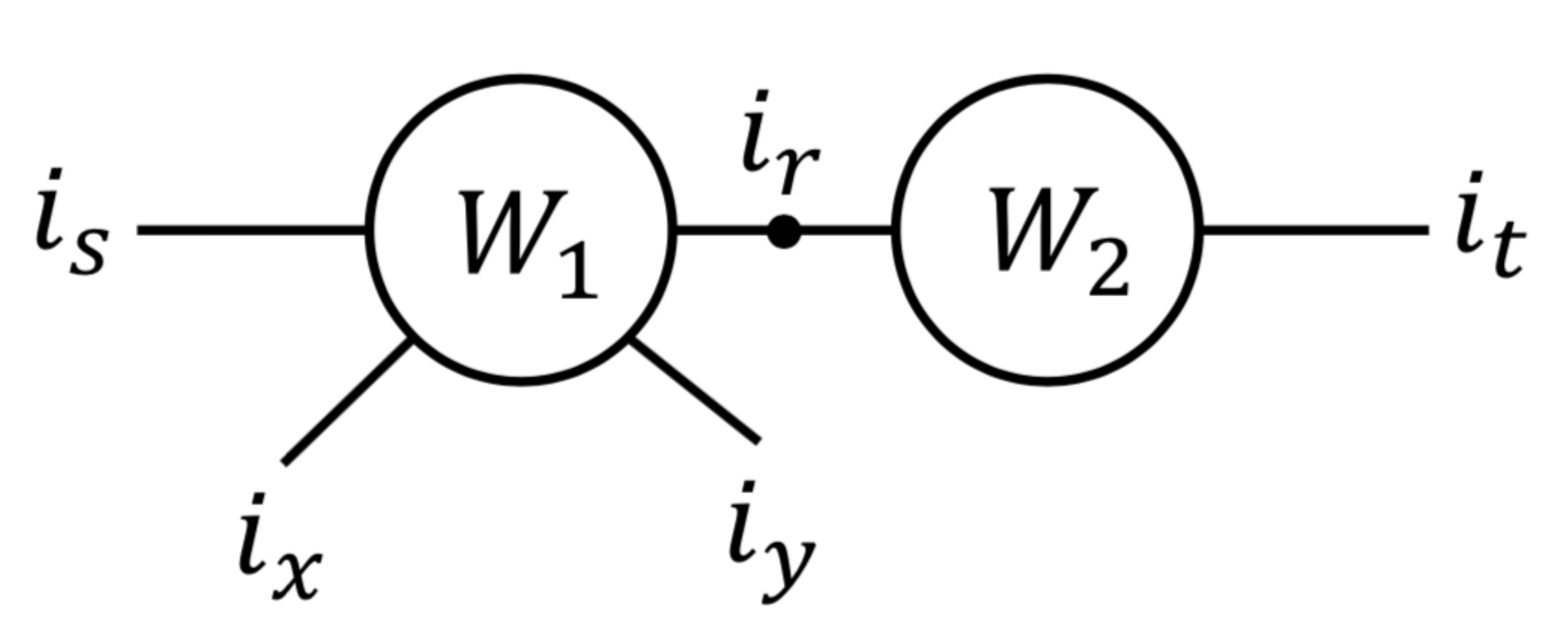}%
  }%
}
&
\vtop{%
\vskip4.5pt
  \hbox{%
    \includegraphics[height=2.2cm]{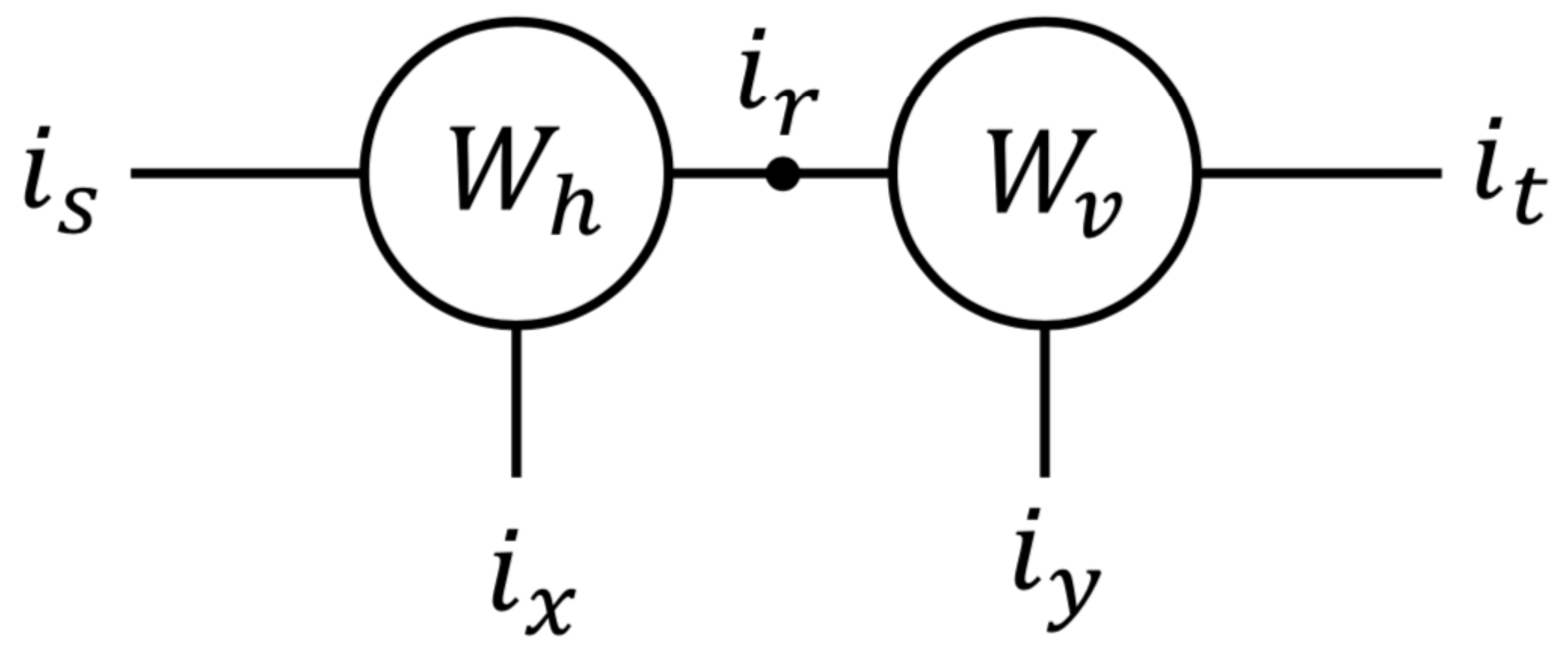}%
  }%
}

\\
(a) & (b) & (c)
\end{tabular}
\end{small}
\caption{Diagram of SVD-based decomposition approaches for a convolutional layer. Each of the nodes represents a factor in the decomposed layer. Edges depict indices of the factors used for the summation. Edges connecting two factors represent pairs of indices used for sum of product operation. (a) Original convolutional layer, (b) weight SVD, (c) spatial SVD.}
\label{fig:learned_weights}
\end{figure*}

The approximate kernel $\widetilde{\mathbf{W}}$ of rank $r$ is expressed as follows:
\begin{equation}
\widetilde{W}(i_x^\prime,i_y^\prime,i_s,i_t) = \sum_{i_r=1}^{r} W_1(i_x^\prime,i_y^\prime,i_s,i_r) W_2(i_r,i_t).
\label{eq:weight_svd}
\end{equation}
The schematic diagram of the summation is given in the figure 2(b). The factors can be obtained using SVD decomposition of $\mathbf{W}=\mathbf{U}\mathbf{S}\mathbf{V}^T$ and assigning $\mathbf{W}_1=\mathbf{U}\mathbf{S}^{\frac{1}{2}}$ and $\mathbf{W}_2=\mathbf{S}^{\frac{1}{2}}\mathbf{V}^T$. The first factor $\mathbf{W}_1$ corresponds to a convolution with a filter size $k\times k$, $s$ input channels and $r$ output channels whereas the second factor corresponds to $1\times 1$ convolution with $r$ input channels and $t$ output channels. The total number of MACs in the decomposed layer equals $c(r)=k^2srhw+rthw$. The compression ratio of the decomposed layer is fully determined by the rank $r$.





\paragraph{Spatial SVD.} This method is based on reshaping the kernel to a matrix $\mathbf{W} \in \mathbb{R}^{sk \times tk}$. The corresponding low-rank approximation of rank $r$ can be expressed as (see figure 2 (c)): 
\begin{equation}
\widetilde{W}(i_s,i_x^\prime,i_t,i_y^\prime) = \sum_{i_r=1}^{r} W_h(i_s,i_x^\prime,i_r) W_v(i_r,i_t,i_y^\prime).
\label{eq:spatial_svd}
\end{equation}
The factor $W_v(i_r,i_t,i_y^\prime)$ corresponds to a convolution with a vertical filter of size $k\times 1$ and the factor $W_h(i_s,i_x^\prime,i_r)$ corresponds to a horizontal $1\times k$ convolution. The total number of MACs is $c(r)=krswh+krtwh$. The trade-off between the computational complexity and approximation error is defined by the rank $r$.

The decomposition was introduced in~\cite{jaderberg2014speeding}. In the original paper, an iterative optimization algorithm based on conjugate gradient descent was used to calculate the factorization. In a subsequent work of ~\cite{tai2015convolutional}, the iterative scheme was replaced by a closed-form solution based on SVD decomposition.

\subsubsection{Tensor decompositions}
\begin{figure*}
\label{fig:conv_tensor_decompositions}
\centering
\begin{small}
\begin{tabular}{c}
\includegraphics[height=3.95cm]{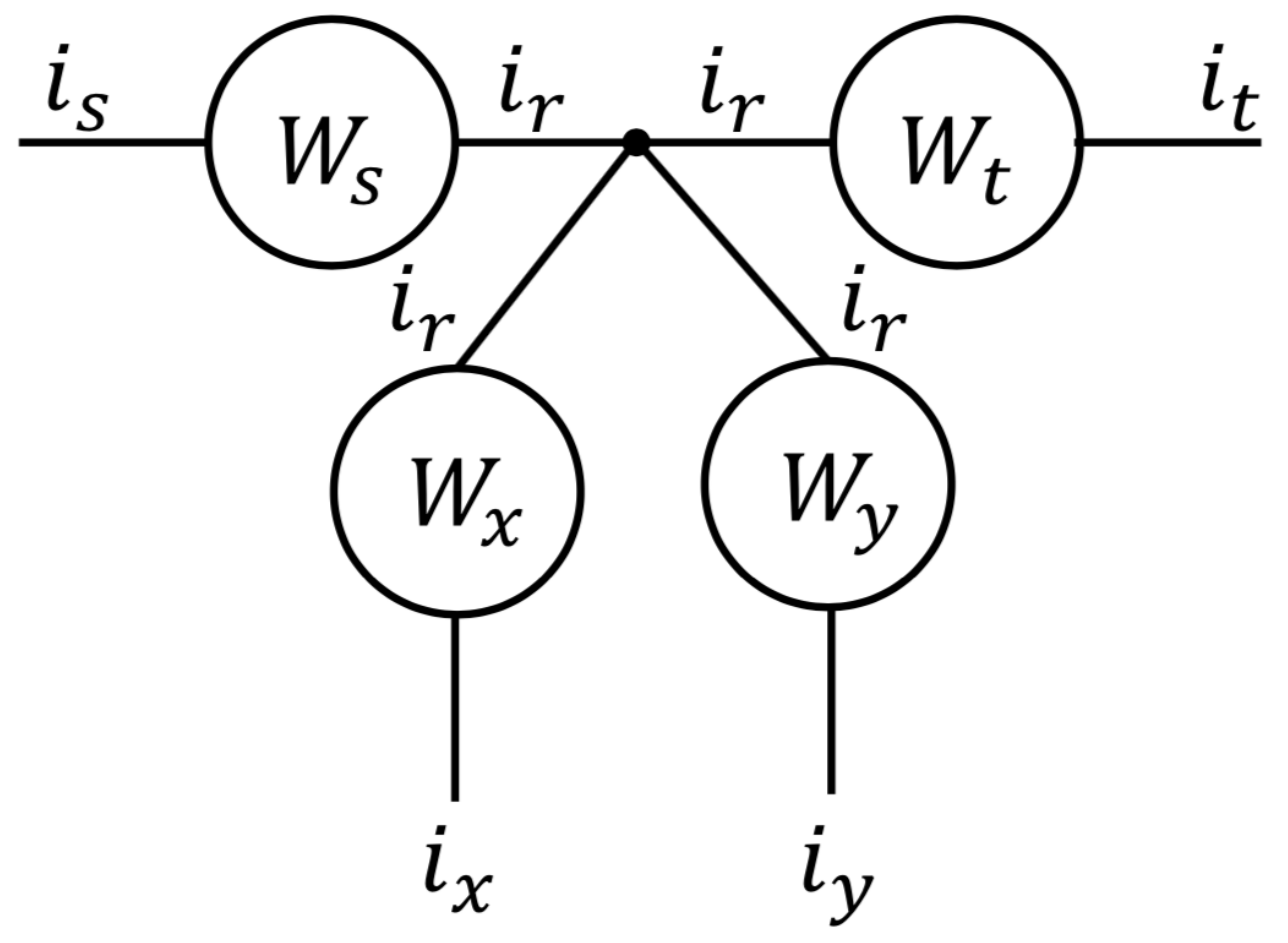} 
\\
(a)
\\
\includegraphics[height=2.5cm]{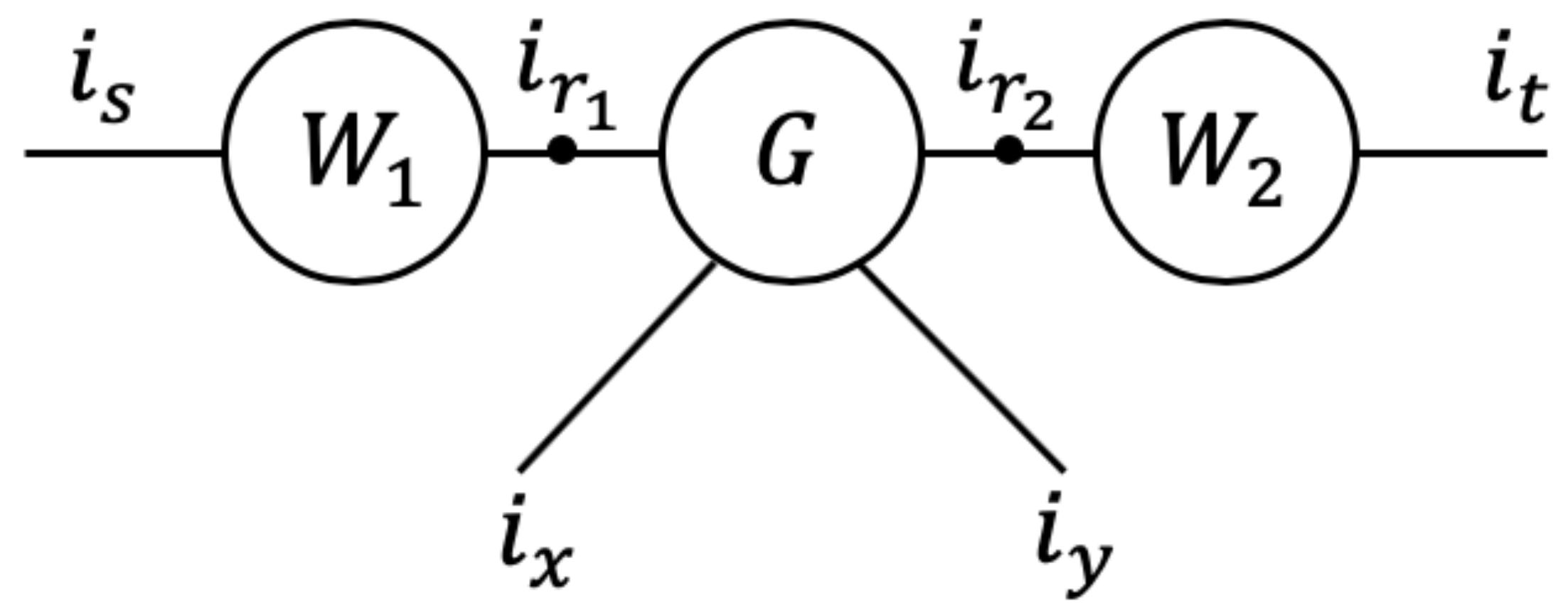} 
\\
(b)
\\
\includegraphics[height=2.9cm]{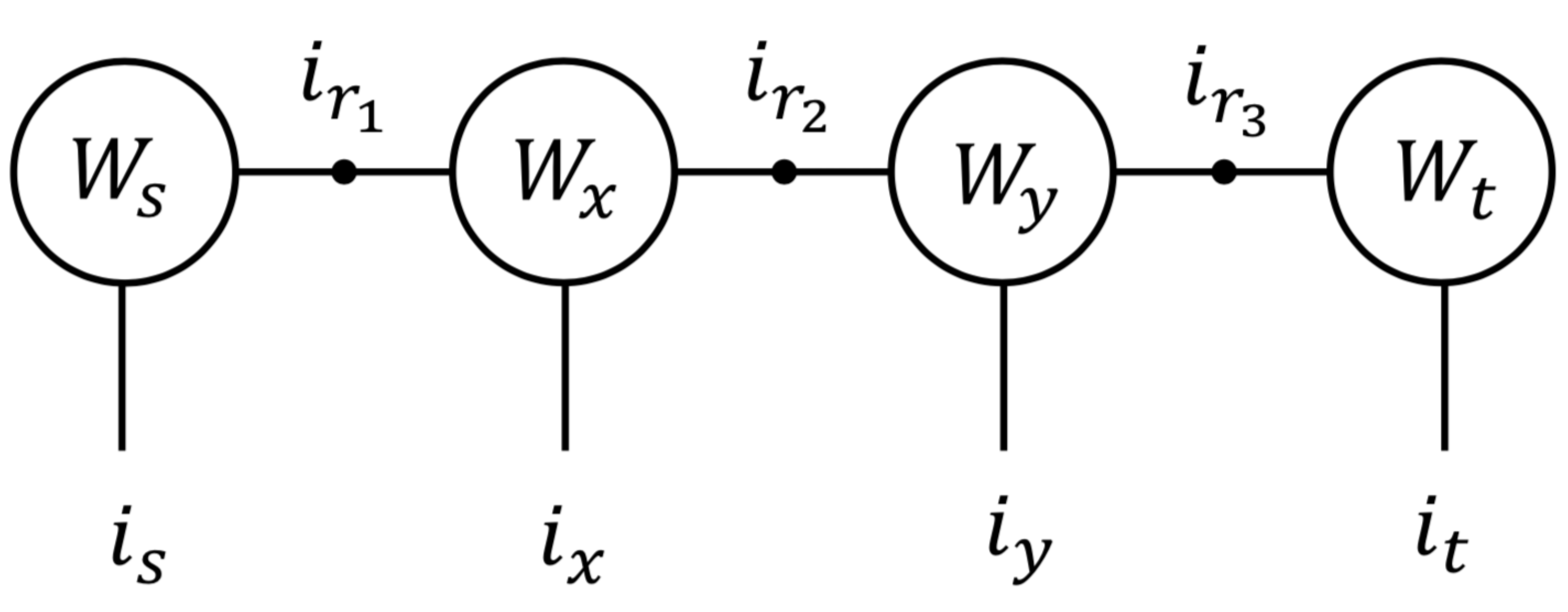}
\\
(c)
\end{tabular}
\end{small}
\caption{Schematic view of the considered tensor decomposition approaches. (a) CP-decomposition. All the four factors  share the same rank $r$ used in the summation in Eqn.~\ref{eq:cp_definition} and have one index for each corresponding to the original indices $i_x^\prime$, $i_y^\prime$, $i_s$, and $i_t$ of the convolution (Eqn.~\ref{eq:conv_orig}).  (b) Tucker decomposition. The core factor $G(i_x^\prime,i_y^\prime,i_{r_1},i_{r_2})$ which can be viewed as a compressed version of the original tensor shares indices $i_{r_1}$ and $i_{r_2}$ with the factors $W_1(i_s,i_{r_1})$ and $W_2(i_t,i_{r_2})$, respectively. (c) Tensor-train decomposition. The original operation is decomposed into a chain of four factors $W^{(1)}(i_s,i_{r_1})$, $W^{(2)}(i_{r_1},i_x^\prime,i_{r_2})$, $W^{(3)}(i_{r_2},i_y^\prime,i_{r_3})$, $W^{(4)}(i_{r_3},i_t)$, each of which keeps one index of the original convolution kernel. Each of the ranks $r_1$, $r_2$, and $r_3$ is shared between a pair of subsequent factors.} 
\end{figure*}

In addition to matricization of the convolutional kernel, several compression techniques based on tensor decompositions were suggested in \cite{lebedev2014speeding,kim2015compression,su2018tensorized} where the kernel is directly treated as a 4-dimensional tensor and decomposed. In this case, different choice of dimensions order in the kernel yields different factorizations. 

\paragraph{CP-decomposition.}
For a kernel $\mathbf{W}\in \mathbb{R}^{s \times k \times k \times t}$, the CP-decomposition of rank $r$ is defined as follows (\cite{kolda2009tensor}):
\begin{equation}
\widetilde{W}(i_s,i_y^\prime,i_x^\prime,i_t)=\sum_{i_r=1}^{r} W^{(s)}(i_s,i_r) W^{(y)}(i_y^\prime,i_r) W^{(x)}(i_x^\prime,i_r)  W^{(t)}(i_t,i_r).  
\label{eq:cp_definition}
\end{equation}
Given the factorization, the original convolution (Eqn.~\ref{eq:conv_orig}) can be approximately computed in four steps~(\cite{lebedev2014speeding}):
\begin{gather}
X^{(1)}(i_r,i_x^\prime,i_y^\prime) = \sum_{i_s=1}^{s} W^{(s)}(i_s,i_r) X(i_x^\prime,i_y^\prime,i_s),\\
X^{(2)}(i_y,i_x^\prime,i_r) = \sum_{i_y^\prime=i_y-\delta}^{i_y+\delta} W^{(y)}(i_y^\prime-i_y+\delta,i_r) X^{(1)}(i_y^\prime,i_x^\prime,i_r),\\
X^{(3)}(i_x,i_y,i_r) = \sum_{i_x^\prime=i_x-\delta}^{i_x+\delta} W^{(x)}(i_x^\prime-i_x+\delta,i_r), X^{(2)}(i_x^\prime,i_y,i_r),\\
Y(i_t,i_x,i_y) = \sum_{i_r=1}^{r} W^{(t)}(i_t,i_r) X^{(3)}(i_x,i_y,i_r),
\end{gather}
where $X^{(1)}(i_r,i_x^\prime,i_y^\prime)$, $X^{(2)}(i_x^\prime,i_y,i_r)$, and $X^{(3)}(i_x,i_y,i_r)$ are tensors which are intermediate results after each step. The diagram of the decomposition and the indices used in the summations is given in the figure 3(a). Computing $X^{(1)}(i_r,i_x^\prime,i_y^\prime)$ and $Y(i_t,i_x,i_y)$ corresponds to convolutions with filter size $1\times 1$. The steps of computing $X^{(2)}(i_x^\prime,i_y,i_r)$ and $X^{(3)}(i_x,i_y,i_r)$ in turn corresponds to convolutions with vertical and horizontal filters, respectively. The total number of MACs in the decomposed layer is $c(r)=(swh + 2kwh + twh)r$. Thus, it depends solely on the value of the rank $r$ which controls the approximation error and computational complexity.    

\paragraph{Tucker decomposition.} For a kernel $W\in \mathbb{R}^{k \times k \times s \times t}$, a partial Tucker decomposition (\cite{kolda2009tensor}) is defined as:
\begin{equation}
\widetilde{W}(i_x^\prime,i_y^\prime,i_s,i_t)=\sum_{i_{r_1}=1}^{r_1}\sum_{i_{r_2}=1}^{r_2} G(i_x^\prime,i_y^\prime,i_{r_1},i_{r_2})W^{(1)}(i_s,i_{r_1}) W^{(2)}(i_t,i_{r_2}), 
\label{eq:tucker_decomposition}
\end{equation}
where $G(i_x,i_y,i_{r_1},i_{r_2})$ is a core tensor of size $k \times k \times r_1 \times r_2$ and $W^{(1)}(i_s,i_{r_1})$ and $W^{(2)}(i_t,i_{r_2})$ are the factor matrices. Computation of the convolution can be decomposed into the following three steps (\cite{kim2015compression}):
\begin{equation}
X^{(1)}(i_{r_1},i_x^\prime,i_y^\prime) = \sum_{i_s=1}^{s} W^{(1)}(i_s,i_{r_1}) X(i_x^\prime,i_y^\prime,i_s),
\label{eq:tucker_first}
\end{equation}
\vspace{-0.0cm}
\begin{equation}
X^{(2)}(i_x,i_y,i_{r_2}) =  \sum_{i_x^\prime=i_x-\delta}^{i_x+\delta} \sum_{i_y^\prime=i_y-\delta}^{i_y+\delta} \sum_{i_{r_1}=1}^{r_1} G(i_x^\prime-i_x+\delta,i_y^\prime-i_y+\delta,i_{r_1},i_{r_2}) X^{(1)} (i_x^\prime,i_y^\prime,i_{r_1}),
\label{eq:tucker_second}
\end{equation}
\vspace{-0.0cm}
\begin{equation}
Y(i_t,i_x,i_y) =  \sum_{i_{r_2}=1}^{r_2} W^{(2)}(i_t,i_{r_2})X^{(2)}(i_x,i_y,i_{r_2}),
\label{eq:tucker_third}
\end{equation}
where the steps in Eqn.~\ref{eq:tucker_first} and Eqn.~\ref{eq:tucker_third} correspond to convolutions with filter size $1\times 1$, and the step in Eqn.~\ref{eq:tucker_second} corresponds to a convolution with the original filter size with $r_1$ input channels and $r_2$ output channels. The total number of MACs is $c(\textbf{r})=sr_1wh+k^2r_1r_2wh+tr_2wh$ and defined by two ranks $r_1$ and $r_2$, so that there is one degree of freedom when selecting the ranks given a predefined compression ratio. The original work by~\cite{kim2015compression} suggests using variational Bayesian matrix factorization~(\cite{nakajima2013global}) for rank selection. 

\paragraph{Tensor-train decomposition.} After reordering the modes as $W\in \mathbb{R}^{s \times k \times k \times t}$, a tensor-train decomposition for the kernel is defined as the following sequence of matrix products (\cite{oseledets2011tensor}): 
\begin{equation}
\begin{split}
\widetilde{W}(i_s,i_x^\prime,i_y^\prime,i_t)= \sum_{i_{r_1}=1}^{r_1}\sum_{i_{r_2}=1}^{r_2}\sum_{i_{r_3}=1}^{r_3}W^{(1)}(i_s,i_{r_1})W^{(2)}(i_{r_1},i_x^\prime,i_{r_2})W^{(3)}(i_{r_2},i_y^\prime,i_{r_3}
)W^{(4)}(i_{r_3},i_t). 
\end{split}
\label{eq:tensor_train_decomposition}
\end{equation}
The original convolution (Eqn.~\ref{eq:conv_orig}) can be computed in four stages~(\cite{su2018tensorized}):
\begin{equation}
X^{(1)}(i_{r_1},i_x^\prime,i_y^\prime) = \sum_{i_s=1}^{s} W^{(1)}(i_s,i_{r_1}) X(i_x^\prime,i_y^\prime,i_s),
\label{eq:tensor_train_first}
\end{equation}
\vspace{-0.0cm}
\begin{equation}
X^{(2)}(i_x,i_y^\prime,i_{r_2}) = \sum_{i_{r_1}=1}^{r_1}\sum_{i_x^\prime=i_x-\delta}^{i_x+\delta} W^{(2)}(i_{r_1},i_x^\prime-i_x+\delta,i_2) X^{(1)}(i_x^\prime,i_y^\prime,i_{r_2}),
\label{eq:tensor_train_second}
\end{equation}
\vspace{-0.0cm}
\begin{equation}
X^{(3)}(i_x,i_y,i_{r_3}) = \sum_{i_{r_2}=1}^{r_2}\sum_{i_y^\prime=i_y-\delta}^{i_y+\delta} W^{(3)}(i_{r_2},i_y^\prime-i_y+\delta,i_{r_3}) X^{(2)}(i_x,i_y^\prime,i_{r_2}),
\label{eq:tensor_train_third}
\end{equation}
\vspace{-0.0cm}
\begin{equation}
Y(i_t,i_x,i_y) = \sum_{r_3=1}^{r_3} W^{(4)}(i_t,i_{r_2}) X^{(3)}(i_x,i_y,i_{r_3}).
\label{eq:tensor_train_fourth}
\end{equation}
The steps in Eqn.~\ref{eq:tensor_train_first}, and Eqn.~\ref{eq:tensor_train_fourth} correspond to $1 \times 1$ convolutions and the steps in Eqn.~\ref{eq:tensor_train_second}, and Eqn.~\ref{eq:tensor_train_third} correspond to a convolution with vertical and horizontal filters, respectively. The total number of MACs is $c(\mathbf{r})=sr_1wh + kr_{1}r_{2}wh + kr_{2}r_{3}wh + r_{3}twh$. The decomposition has three ranks $r_1$, $r_2$, and $r_3$ that determine the approximation error and the computational complexity of the compressed layer.


\begin{center}
\begin{table*}
\begin{normalsize}
  \centering
  \begin{tabular}{| c |c | c | c|}
    \hline
    & Num. parameters & Comp. complexity & \parbox[height=1.5cm]{1.1cm}{Num. \\ ranks} \\ \hline
    Original & $k^2st$ & $k^2whst$ & -\\ \hline
    Weight SVD & $(k^2s+t)r$ & $(k^2whs+wht)r$ & 1\\ \hline
    Spatial SVD & $(ksr+kt)r$ & $(kwhs+kwht)r$ & 1 \\ \hline
    CP decomposition & $(2k+s+t)r$ & $(swh + 2kwh + twh)r$  & 1\\ \hline
    Tucker decomposition  & $sr_1+k^2r_1r_2+tr_2$ & $sr_1wh+k^2r_1r_2wh+tr_2wh$ & 2 \\ \hline
    Tensor-train  & $sr_1+kr_1r_2+ kr_2r_3+r_3t$ & \parbox{3.5cm}{$sr_1wh + kr_1r_2wh+$ \\ $kr_2r_3wh + r_3twh$} & 3\\ \hline
    \end{tabular}
\caption{Comparison of SVD and tensor decomposition methods in terms of computational complexity and the number of parameters.}
\label{tab:runtime}
\end{normalsize}
\end{table*}
\end{center}


\begin{figure}[t]
\centering
\begin{small}
\begin{tabular}{c}
\includegraphics[scale=0.2]{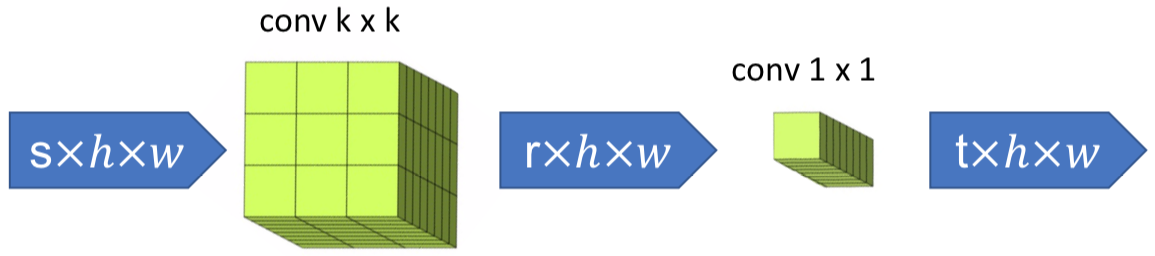}\\
\vspace{0.3cm}(a) Weight SVD\vspace{0.3cm} \\
\includegraphics[scale=0.2]{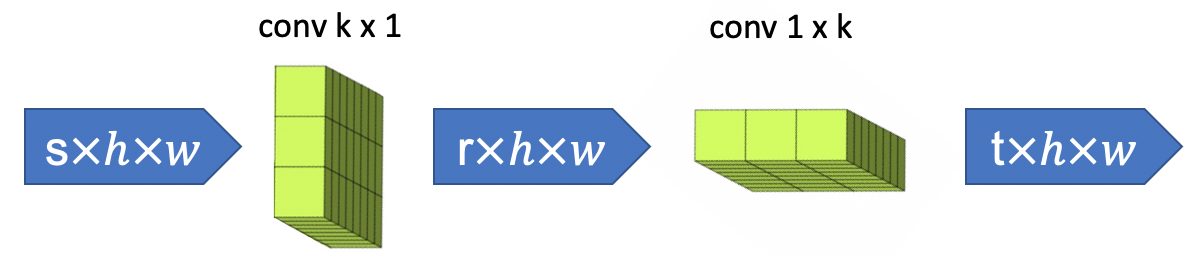}\\
\vspace{0.1cm}(b) Spatial SVD\vspace{0.3cm} \\
\includegraphics[scale=0.2]{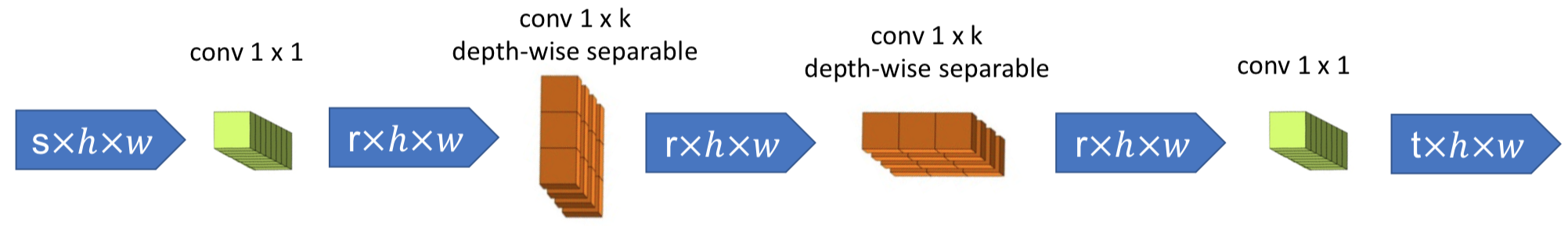}\\
\vspace{0.3cm}(c) CP-decomposition\vspace{0.3cm} \\
\includegraphics[scale=0.2]{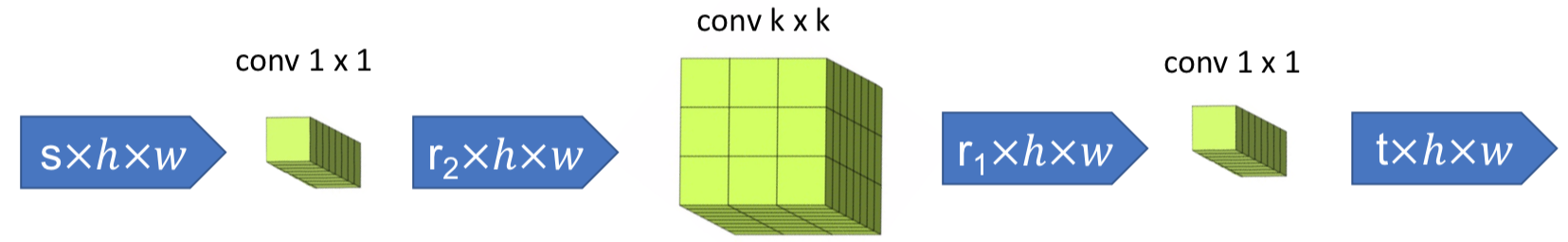}\\
\vspace{0.3cm}(d) Tucker decomposition \vspace{0.3cm}\\
\includegraphics[scale=0.2]{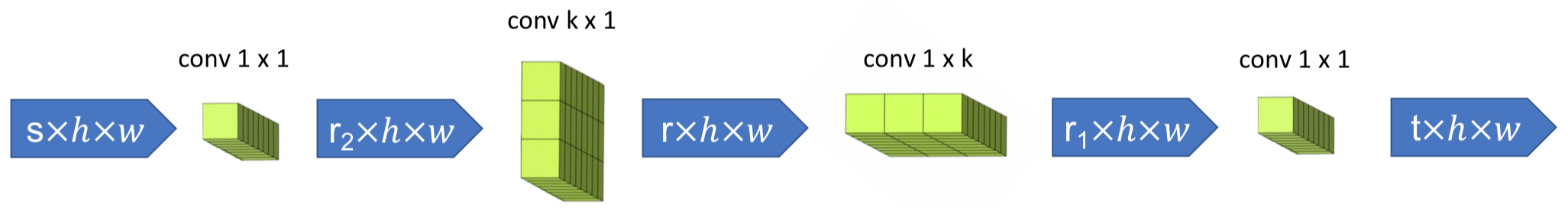} \\
\vspace{0.3cm}(e) Tensor-train decomposition \vspace{0.3cm}
\label{fig:methodology_levels}
\end{tabular}
\end{small}
\caption{Overview of different tensor decomposition approaches.}

\end{figure}

%% file: sections/methods_level_2.tex
\subsection{Level 2. Data driven compression methods}
\label{section:level_2}
\subsubsection{Per-layer data-optimized SVD methods}

All level 1 compression methods minimize the kernel approximation error. This does not use any information of the actual data which is being processed by the layer of the network. One of the ways to improve level 1 methods is to formulate a method that minimizes the error in the activations produced by the compressed layer. A method based on minimizing the error of the output for the specific data allows one to significantly decrease the loss in accuracy after compression. This section describes multiple approaches for per-layer data-optimized SVD compression.

\noindent\textbf{Data SVD.} Given a kernel tensor $\mathbf{W}$ reshaped into a matrix of shape $t \times k^2s$, an input vector $x \in \mathbb{R}^{k^2s}$, the response $y \in \mathbb{R}^{t}$ is given by:
\begin{equation}
\mathbf{y} = \mathbf{W}\mathbf{x}+\mathbf{b}. \end{equation}
Given the output data, the optimal projection matrix $\mathbf{M} \in \mathbb{R}^{t\times t}$ is given as a solution of the following optimization problem:
\begin{equation}
\begin{aligned}
& \underset{\mathbf{M}}{\text{argmin}}
& & \sum_{i=1}^{n}\norm{(\mathbf{y_i}-\mathbf{\overline{y}})-\mathbf{M}(\mathbf{y_i}-\mathbf{\overline{y}})}^2_2 \\
& \text{s.t.}
& & \rank \mathbf{M} \leq r,
\end{aligned}
\label{eq:data_svd}
\end{equation}
where $\mathbf{y_i}$ are outputs sampled from the training set, $\mathbf{\overline{y}}$ is the sample mean, and $n$ is the number of samples. The solution is given by principal component analysis (PCA) as follows~\citep{golub1996matrix}. Let $\mathbf{\widehat{Y}}\in \mathbb{R}^{t\times n}$ be a matrix which concatenates the entries of $(\mathbf{y_i}-\mathbf{\overline{y}})$. Given the eigendecomposition of the covariance matrix $\mathbf{\widehat{Y}}\mathbf{\widehat{Y}}^T=\mathbf{U}\mathbf{S}\mathbf{U}^T$, the values of $\mathbf{M}$ are given by:
\begin{equation}
\mathbf{M}=\mathbf{U}_{r}\mathbf{U}_{r}^T,   
\label{eq:data_svd_low_rank}
\end{equation}
where $\mathbf{U}_{r}$ are the first $r$ are eigenvectors. 
This solution for $\mathbf{M}$ can be used to approximate the original layer. Under the low rank assumption for vector $\mathbf{y}$, the output can be expressed as:
\begin{equation}
\mathbf{y}=\mathbf{M}\mathbf{W}\mathbf{x} + \mathbf{b}, 
\end{equation}
where $\mathbf{x}$ is the input vector and $\mathbf{b}$ is the bias. Using Eqn.~\ref{eq:data_svd_low_rank}, the original kernel $\mathbf{W}$ can be approximated as $\mathbf{\widetilde{W}}=\mathbf{W_1}\mathbf{W_2}$, where $\mathbf{W}_1=\mathbf{U}_{R}$, and $\mathbf{W}_2=\mathbf{U}_{R}^T\mathbf{W}$. This method corresponds to a data-optimized version of the weight SVD decomposition (Eqn.~\ref{eq:weight_svd}).

\noindent\textbf{Asymmetric data SVD.}
One of the main issues in neural network compression is the accumulation of error when compressing a deep model. Since every layer is compressed subsequently, compressed layers could take into account the error introduced by previous layers in their decomposition for better performance.  An asymmetric formulation was introduced to do this in~\cite{zhang2016accelerating}. As opposed to optimizing the reconstruction error for the approximated layer based on the original input data, the asymmetric formulation is based on the input data from the previous approximated layer. This approach allows one to significantly reduce the full-model accuracy drop in level 2 settings using a limited amount of input data at the cost of solving a more general optimization problem. 

Given the output of the previous compressed layer $\widehat{\mathbf{x}}$, the activations are given by:
\begin{equation}
\mathbf{z} = \mathbf{W}\widehat{\mathbf{x}}+\mathbf{b}.   
\end{equation}
In order to minimize the error introduced by compression, the following optimization problem is solved:
\begin{equation}
\begin{aligned}
& \underset{\mathbf{M}}{\text{argmin}}
& & \sum_{i=1}^{n}\norm{(\mathbf{y_i}-\mathbf{\overline{y}})-\mathbf{M}(\mathbf{z_i}-\mathbf{\overline{z}})}^2_2 \\
& \text{s.t.}
& & \rank \mathbf{M} \leq r.
\end{aligned}
\label{eq:data_svd_asym}
\end{equation}
The problem is based on minimizing the same error as in Eqn.~\ref{eq:data_svd}, but it depends on both the original layer outputs $\mathbf{y_i}$, and the compressed layer outputs $\mathbf{z_i}$. After combining responses $(\mathbf{y_i}-\mathbf{\overline{y}})$, and $(\mathbf{z_i}-\mathbf{\overline{z}})$ into matrices $\mathbf{\widehat{Y}}$ and $\mathbf{\widehat{Z}}$, the minimization can be written as:
\begin{equation}
\begin{aligned}
& \underset{\mathbf{M}}{\text{argmin}}
& & \norm{\mathbf{\widehat{Y}}-\mathbf{M}\mathbf{\widehat{Z}}}^2_F \\
& \text{s.t.}
& & \rank \mathbf{M} \leq r.
\end{aligned}
\label{eq:opt_gsvd}
\end{equation}
The problem has a closed form solution for $M$ based on generalized SVD~\citep{takane2006generalized}. The new bias for the compressed layer can be computed as $\mathbf{b}_{new}=\overline{\mathbf{z}}-\mathbf{M}\overline{\mathbf{y}}$.

The reconstruction error of the asymmetric data SVD can be further improved be incorporating the activation function into the formulation in Eqn.~\ref{eq:opt_gsvd}.
\begin{equation}
\begin{aligned}
& \underset{\mathbf{M},\mathbf{\widehat{b}}}{\text{argmin}}
& & \norm{f(\mathbf{Y})-f(\mathbf{M}\mathbf{\widehat{Z}}+\mathbf{\widehat{b}})}^2_F \\
& \text{s.t.}
& & \rank \mathbf{M} \leq r,
\end{aligned}
\label{eq:asym_nonlin}
\end{equation}
Where $\mathbf{Y}$ is a matrix concatenating the entries of $\mathbf{y_i}$ with no mean subtracted, and $\widehat{\mathbf{b}}$ is a new bias. This problem is solved using the following relaxation:
\begin{equation}
\begin{aligned}
& \underset{\mathbf{M},\widehat{\mathbf{b}},\mathbf{Z}}{\text{argmin}}
& & \norm{f(\mathbf{Y})-f(\mathbf{Z})}^2_F + \lambda \norm{\mathbf{Z}-\mathbf{M}\mathbf{\widetilde{Z}}-\mathbf{\widehat{b}}}^2_F \\
& \text{s.t.}
& & \rank \mathbf{M} \leq r,
\label{eq:asym_nonlin_relaxed}
\end{aligned}
\end{equation}
where $\mathbf{Z}$ is an auxiliary variable, and $\lambda$ is a penalty parameter. The second term of the objective is equivalent to Eqn.~\ref{eq:opt_gsvd}. The first term can be minimized using SGD for any activation function, or in the case of the ReLU function, it can be solved analytically (\cite{zhang2016accelerating}). Minimization of  the objective Eqn.~\ref{eq:asym_nonlin_relaxed} is performed by using alternating minimization. The first sub-problem corresponds to fixing $\mathbf{\widetilde{Z}}$ and solving for $\mathbf{M}$, $\mathbf{\widehat{b}}$, and vice versa for the second sub-problem. Increasing values of parameters $\lambda$ are used through the iterations of the method.

\noindent\textbf{Asym3D.}
The authors of ~\cite{zhang2016accelerating} further propose to use the formulation in Eqn.~\ref{eq:opt_gsvd} to perform a double decomposition based on the spatial and data SVD methods. Given two spatial SVD layers $\mathbf{W}_v$, $\mathbf{W}_h$, the formulation in Eqn.~\ref{eq:opt_gsvd} can be applied in order to perform a further decomposition of the second layer $\mathbf{W}_v$.
The trade-off between accuracy and the computational complexity in this case is determined by two ranks: $r_s$ is the rank of the original spatial SVD decomposition and rank $r_d$ is the rank of the data optimized decomposition applied to the factor $\mathbf{W}_h$. The final decomposed architecture consists of a $k \times 1$ filter with $r_s$ output channels followed by a $1 \times k$ filter with $r_d$ output channels and a $1\times 1$ convolutional layer with $t$ output channels.

\noindent \textbf{Data optimized spatial SVD.}
In addition to Asym3D method, the framework for per-layer optimization (Eqn.~\ref{eq:data_svd_asym}) can be used to obtain a data-optimized version of the spatial SVD method. If we consider the optimization problem in Eqn.~\ref{eq:data_svd_asym} without the constraint on the rank:
\begin{equation}
\begin{aligned}
& \underset{\mathbf{M}}{\text{argmin}}
& & \sum_{i=1}^{n}\norm{(\mathbf{y_i}-\mathbf{\overline{y}})-\mathbf{M}(\mathbf{z_i}-\mathbf{\overline{z}})}^2_2, 
\end{aligned}
\label{eq:data_svd_asym_full_rank}
\end{equation}
the solution for $\mathbf{M}$ can be used to improve the predictions by refining weights of a compressed network layer based on some input and output data.
Consider a convolutional layer decomposed using the spatial SVD decomposition (Eqn.~\ref{eq:spatial_svd}). Given the original weights $\mathbf{W}$, the layer can be decomposed into two layers:
\begin{equation}
\mathbf{W}=\mathbf{W}_v \mathbf{W}_h.    
\label{eq:gsvd_spatial}
\end{equation}
Given an input vector $\widehat{\mathbf{x}}$, the output ${\mathbf{z}}$ is given by:
\begin{equation}
\mathbf{z}=\mathbf{W}_v \mathbf{W}_h \mathbf{\widehat{x}}.    
\end{equation}
After solving Eqn.~\ref{eq:data_svd_asym_full_rank} for $\mathbf{z}$ above and the reference output, the data-optimized version of the weights $\widetilde{\mathbf{W}}$ is given as:
\begin{equation}
\widetilde{\mathbf{W}}=\mathbf{M}\mathbf{W}=(\mathbf{M} \mathbf{W}_v) \mathbf{W}_h.    
\label{eq:gsvd_spatial_opt}
\end{equation}
In practice, the refined value $\widetilde{\mathbf{W}}_v=\mathbf{M} \mathbf{W}_v$ can be used instead $\mathbf{W}_v$ for the second layer.


\subsubsection{Channel pruning}
Some compression methods introduced in the literature are based on pruning channels of a convolutional filter based on different channel importance criteria. In particular, the method suggested in~\cite{li2016pruning} is based on the weight magnitudes. Another pruning method which is optimized for data was introduced in~\cite{he2017channel}. This method uses lasso feature selection to find the set of channels to prune. 

In order to formulate the pruning method as an optimization, the authors consider computing the output of a convolutional layer with a kernel $\mathbf{W}\in \mathbb{R}^{t \times s \times k \times k}$ on input volumes $\mathbf{X}\in \mathbb{R}^{n \times s \times k \times k}$ sampled from the feature map of the uncompressed model, where $n$ is the number of samples. The corresponding output volume $\mathbf{Y}$ is a matrix of shape $n \times t$. The original number of channels is reduced to $s^\prime$ ($0 \leq s^\prime \leq s$) in a way that the reconstruction error for the output volume is minimized. The objective function is:
\begin{equation}
\begin{aligned}
& \underset{\beta, \mathbf{W}}{\text{argmin}}
& & \norm{\mathbf{Y}-\sum_{i=1}^{s}\beta_i \mathbf{X}_i \mathbf{W}_i^T}^2_F \\
& \text{s.t.}
& & \norm{\mathbf{\beta}}_0 \leq s',
\end{aligned}
\end{equation}
where $\mathbf{X}_i \in \mathbb{R}^{n \times k^2}$ is an i-th channel of the input concatenated for multiple data samples, and $\mathbf{W}_i \in \mathbb{R}^{t \times k^2}$ is i-th channel of the filter, both are reshaped into matrices. Vector $\mathbf{\beta}$ is the coefficient vector for channel selection. If the value $\beta_i=0$ then the corresponding channel can be pruned. In order to solve the problem, the $L_0$ norm is relaxed to $L_1$ and the minimization is formulated as follows:
\begin{equation}
\begin{aligned}
& \underset{\beta, \mathbf{W}}{\text{argmin}}
& & \norm{\mathbf{Y}-\sum_{i=1}^{s}\beta_i \mathbf{X}_i \mathbf{W}_i^T}^2_F + \lambda \norm{\mathbf{\beta}}_1 \\
& \text{s.t.}
& & \norm{\mathbf{W}_i}_F = 1.
\end{aligned}
\label{eq:channel_pruning}
\end{equation}
The minimization is performed in two steps by fixing $\mathbf{\beta}$ or $\mathbf{W}$ and solving the corresponding sub-problems.

%% file: sections/methods_level_3.tex
\subsection{Level 3. Compression based on training}

Some compression methods require full training of the model. Either by fine-tuning an already trained model for a few training epochs, or training the model entirely from scratch. All of the procedures in the previous paragraphs can be extended this way into an iterative compression and fine-tuning scheme. Here we focus on probabilistic compression methods that need fine-tuning or training from scratch.

\subsubsection{Probabilistic compression}

Several methods have been proposed in the literature that add a, potentially probabilistic, multiplicative factor $z$ to each channel in the convolutional network. Such that we have for a single layer with input x, weight matrix W and output y:
\begin{equation}
y = z(\alpha) \cdot W * x,
\end{equation}
with $z$ the same dimensionality as the output $y$, and $\alpha$ one or more learnable parameters that control the gate. The idea is that when $z$ equals 0, the output channel is off and can be removed from the network. The factor $z$ can also be interpreted as a gate that is on or off. Similarly, in the probabilistic setting, if the gate is sampled close to 0 with a high likelihood or has a very high variance, the channel can be removed. This multiplicative factor is regularized by a penalty term in the loss function, such that during training the network optimizes for the trade-off between the loss function and the model complexity as follows:
\begin{equation}
    \mathcal{\hat{L}}(\mathbf{X}, \mathbf{Y}) = \mathcal{L}(\mathbf{X}, \mathbf{Y}, \alpha) + \lambda F(\alpha),
\end{equation}
where $\mathcal{L}$ is the original loss function, $F$ a differentiable function of the complexity of the network, parametrized by (learnable) parameters $\alpha$ that control the gates, and $\lambda$ a trade-off factor between the two loss functions. In all methods, $\lambda$ is a hyperparameter that is set by the user.

\paragraph{$L_0$-regularization.}
The technique from \citet{christosl0} applies the $L_0$-norm to channels in the neural network. The $L_0$-norm is defined as $\norm{\mathbf{\theta}}_0$, the amount of non-zero entries in a vector $\mathbf{\theta}$. Generally, this norm cannot be optimized directly, but the paper extends the continuous relaxation trick from \citet{concrete, gumbelsoftmax} to optimize the gates.  \citet{christosl0} introduces the hard-concrete distribution for the gate, which is a clipped version of the concrete distribution:
\begin{align}
\mathbf{u} & \sim \mathcal{U}(0, 1), \\
\mathbf{s}& = \textrm{Sigmoid}((\log (\mathbf{u}) - \log(1-\mathbf{u}) + \log(\mathbf{\alpha}))/ \beta), \\  \label{gumbsoftmax}
\bar{\mathbf{s}}& = \mathbf{s}(\zeta - \gamma) + \gamma, \\ 
\mathbf{z}& = \min(1, \max(0, \bar{\mathbf{s}}))
\end{align}

Each channel in a convolutional network is multiplied with one gate $z$, which is parametrized by parameter $\alpha$. In the forward pass, a sample is drawn from the hard-concrete distribution for each gate, creating a stochastic optimization procedure. $\beta$ is the temperature parameter, set as $\beta=2/3$ in the paper, which controls the skew of the sigmoid. Parameters $\zeta, \gamma$ are stretching factors for clipping some values to actual 0s and 1s, which are set to $1.1$ and $-0.1$, respectively.
The method penalizes the probability that each gate is sampled as $1$. Channels corresponding to gates that have a low probability of being active can be removed from the network. This corresponds to a small parameter $\alpha$. The regularization factor chosen here is:
\begin{equation}
    \mathcal{F}(\alpha) = \sum_{j=1}^{N_g} \textrm{Sigmoid}(\log(\alpha_j) - \beta \log (\frac{-\gamma}{\zeta})),
\end{equation}
where $N_g$ is the total number of gates in the network.

\paragraph{Variational Information Bottleneck.}
\citet{vibnet} introduces a Gaussian gate that is multiplied with each channel in the network. In the forward pass, a sample is drawn from the Gaussian $\mathcal{N}(\mu, \sigma)$ by using the reparametrization trick from \cite{reparamtrick}. This corresponds to gates $z$ such that:
\begin{align}
    \epsilon& \sim \mathcal{N}(0, 1), \\
    z& = \mu + \epsilon \cdot \sigma,
\end{align}
where $\mu$ and $\sigma$ are learnable parameters, corresponding to the mean and standard deviation of the Gaussian. The corresponding regularization factor is derived to be 

\begin{align}
    \mathcal{F}(\mu, \sigma) = \sum_{j=1}^{N_g} \log \left(1 + \frac{\mu_j^2}{\sigma^2_j} \right),
\end{align}
where again, $N_g$ is the number of gates in the network. The channels that have a small ratio $\mu_j^2/\sigma_h^{2}$ can be removed from the network, as they are either multiplied with a small mean value or have a very large variance. 

The methods from \citet{bayesiancompression} and \citet{neklyudov} are variants of this method with different regularization functions $\mathcal{F}$.

%% file: sections/compression_ratio_selection.tex
\subsection{Compression ratio selection for whole-model compression}
Per layer compression ratio selection is one of the important aspects of neural network compression. In this section we introduce two different methods for compression ratio selection which we used for our experiments.
\label{sec:compression_ratio_selection}
\subsubsection{Equal accuracy loss}
To compare different SVD and tensor decomposition based compression techniques in similar settings, we suggest using the following ratio selection method. The main advantage of this method is that it can be defined for any decomposition approach in a similar way. 

To introduce the rank selection method, we first define a layer-wise accuracy metric based on a verification set. The verification set is a subset of the training set used for the rank selection method to avoid using the validation set. For a layer $l$, the accuracy $P_l(\mathbf{r})$ is obtained by compressing the layer $l$ using a vector of ranks $\mathbf{r}$, while the rest of the networks remains uncompressed. The network with the single compressed layer is evaluated on the verification set to calculate the value $P_l(\mathbf{r})$. In order to avoid extra computational overhead, in practice the layer-wise accuracy metric is calculated only for some values of $\mathbf{r}$, e.g., values of $\mathbf{r}$ that correspond to per-layer compression ratios $\{0.1, 0.2, \dots, 0.9\}$. 






We denote the combination of all rank values for all the layers as $\mathbf{R}=(\mathbf{r}_1,\dots,\mathbf{r}_L)^T$, where each rank $\mathbf{r}_i$ is a scalar in case of SVD decomposition, and a vector in case of high-dimensional tensor decomposition techniques. The set of ranks $\mathbf{R}$ can be calculated as the solution to the following optimization problem. The input consists of per-layer accuracy-based metric $P_l(\mathbf{r}_l)$, the full compressed model complexity $C(\mathbf{R})=\sum_{l=1}^L c_l(\mathbf{r}_l)$, the original model accuracy $P_{orig}$ and the original model complexity $C_{orig}$:
\begin{equation}
\begin{split}
\mathbf{R} = (\mathbf{r}_1,\dots,\mathbf{r}_L)^T = & \underset{\mathbf{Z}=(\mathbf{z}_1,\dots,\mathbf{z}_L)}{\text{argmin}} \tau \\
& \text{ s.t. } P_l(\mathbf{z}_l)\geq P_{orig} - \tau, \\
& \hspace{0.8cm} \frac{C(\mathbf{\mathbf{Z}})}{C_{orig}} \leq \alpha,
\end{split}
\label{eq:rank_selection_equal_acc_loss}
\end{equation}where $\tau$ is the tolerance in per-layer accuracy metric decrease. The tolerance value is iteratively adjusted to meet the desired full model compression ratio $\alpha$.

\subsubsection{Greedy algorithm based on singular values}

To facilitate comparison of data-optimized SVD methods, we use the following method introduced in~\cite{zhang2016accelerating}. The method is based on the assumption that the whole-model performance is related to the following PCA energy:
\begin{equation}
\mathcal{E(\mathbf{R})} = \prod_{l=1}^{L} \sum_{k=1}^{r_l}\sigma_{l,k},
\end{equation}
where $\sigma_{l,:}$ are the singular values of layer $l$. To choose the ranks for SVD decomposition, the energy is maximized subject to the constraint on the total number of MACs in the compressed model:
\begin{equation}
\begin{aligned}
& {\text{max } \mathcal{E(\mathbf{R})}}
& &  \\
& \text{ s.t. } \frac{C(\mathbf{\mathbf{R}})}{C_{orig}} \leq \alpha.
\end{aligned}
\label{eq:channel_pruning}
\end{equation}
To optimize the objective, the greedy strategy of \cite{zhang2016accelerating} is used. This approach has a relatively low computational cost and does not require using the validation set.

%% file: sections/experiments.tex
\section{Experiments}

To evaluate the performance of different compression techniques at different levels, we used a set of the models from PyTorch~(\cite{pytorch}) model zoo, including Resnet18, Resnet50, VGG16, InceptionV3, and MobileNetV2 trained on ImageNet data set. For every model we used 1.33x, 2x, 3x, and 4x compression ratios in terms of MACs, which serves as a proxy for run-time. 

\subsection{Level 1 compression}
\begin{center}
\begin{figure}[t]
\centering
\begin{small}
\begin{tabular}{cc}
\hspace{-0.3cm}\vspace{-0.0cm}\includegraphics[width=7.5cm]{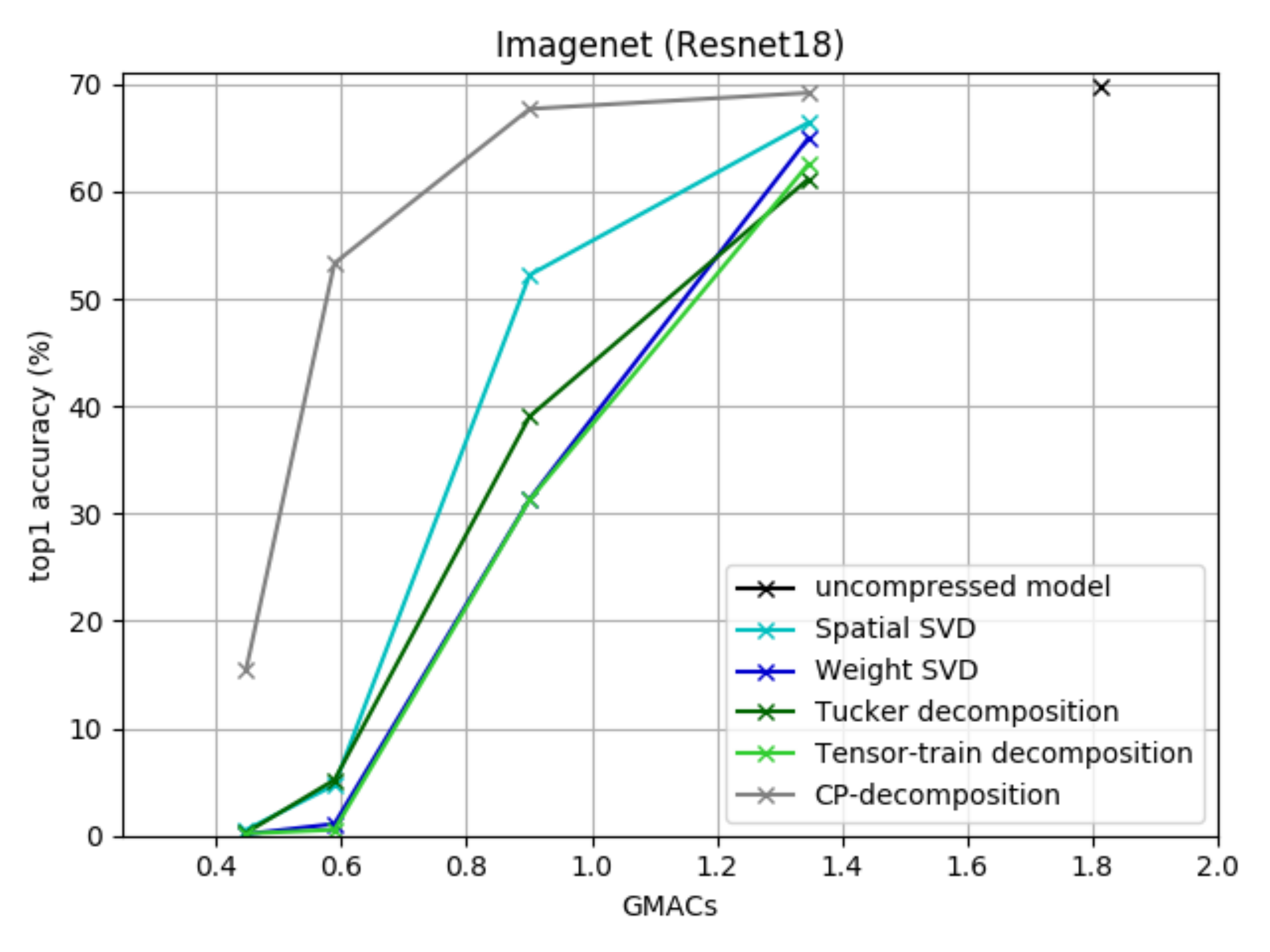}\hspace{-0.0cm} &
\includegraphics[width=7.5cm]{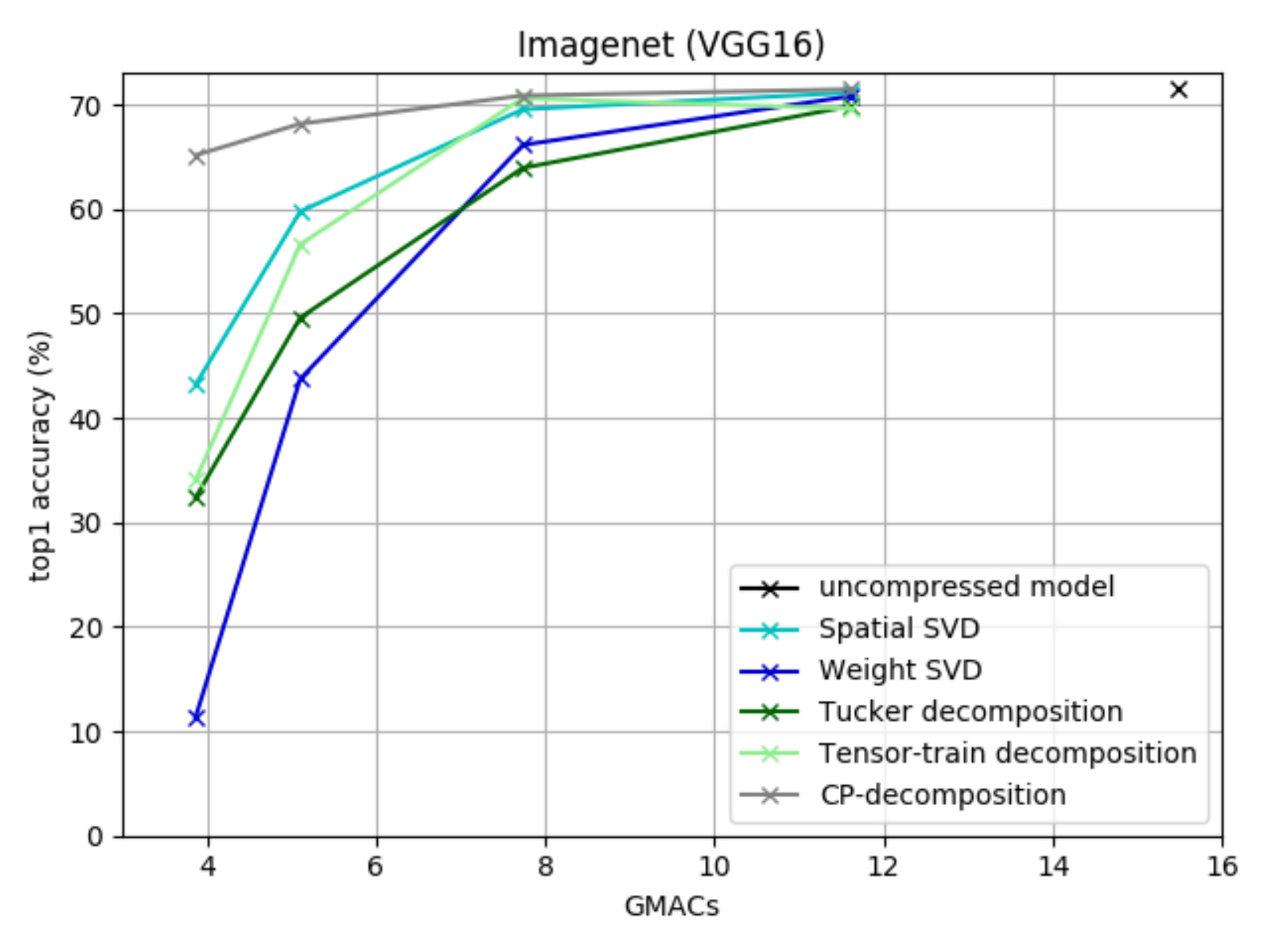}\\
\hspace{-0.3cm}\includegraphics[width=7.5cm]{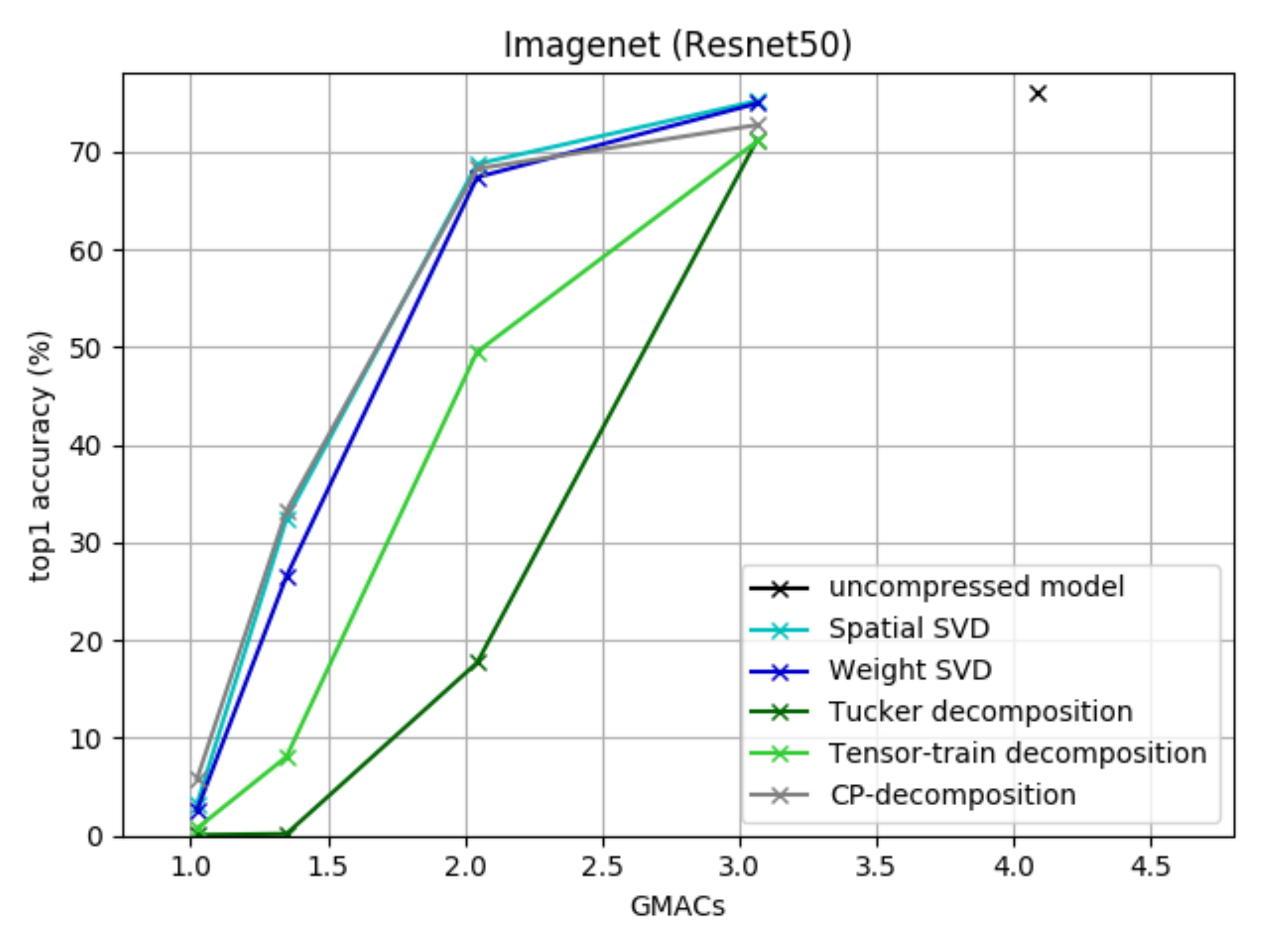}\hspace{-0.0cm} &
\includegraphics[width=7.5cm]{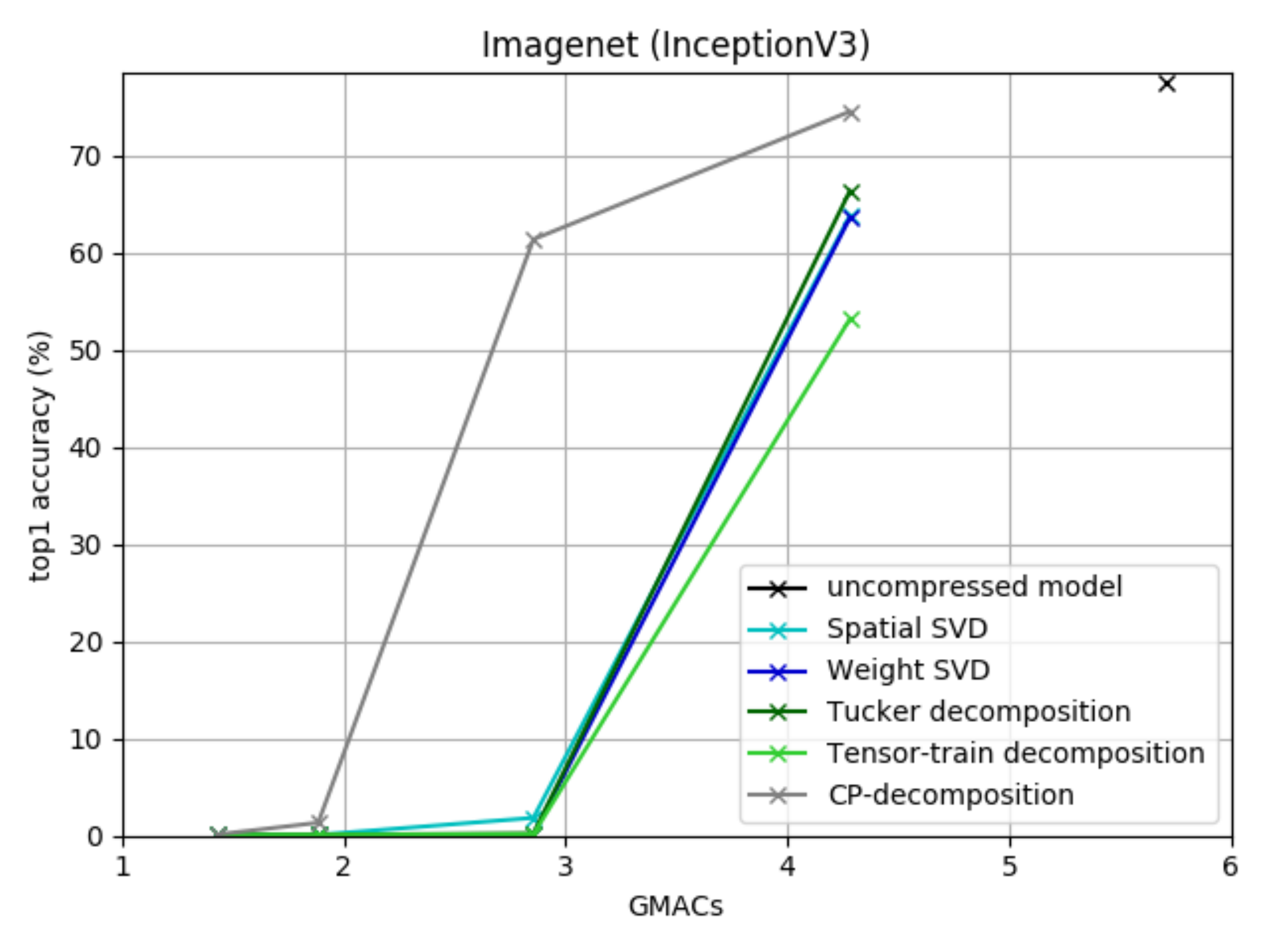}
\end{tabular}
\end{small}
\caption{Level 1 compression. Comparison of different SVD and tensor decomposition methods for Resnet18, VGG16, Resnet50, and InceptionV3 pre-trained on ImageNet. Overall, the best performance is mostly achieved using CP-decomposition for every model. For the greatest part of the experiments, the second best method is the spatial SVD decomposition. The ranking of the other methods depends on the model.}
\label{fig:level_1}
\end{figure}
\end{center}

To compare the performance of level 1 compression techniques, we used Resnet18, VGG16, Resnet50, and InceptionV3 models, no fine-tuning or data-aware optimization was used. Five different compression techniques were evaluated, including spatial SVD, weight SVD, Tucker decomposition, tensor-train decomposition, and CP-decomposition. To compute compression ratios per layer, the method based on equal accuracy loss was used for all the methods.

For decomposition approaches that have a single rank value including spatial SVD, weight SVD, and CP-decomposition, the rank value is fully determined by the compression ratio. For Tucker decomposition, we add an additional constraint $\frac{r_1}{\widehat{r}_{1}} \approx \frac{r_2} {\widehat{r}_{2}}$ to calculate the ranks, where $\widehat{r}_1$ and $\widehat{r}_2$ are the maximum values of ranks $r_1$ and $r_2$ respectively (definition of ranks for Tucker decomposition is given in Eqn.~\ref{eq:tucker_decomposition}), and the equality is approximate due to integer values of the ranks.
In a similar way, for tensor-train decomposition we use the pair of constraints $\frac{r_1}{\widehat{r_1}} \approx \frac{r_2} {\widehat{r}_{2}} \approx \frac{r_3} {\widehat{r}_{3}}$ to determine the set of three ranks based on the compression ratio value, where $\widehat{r}_1$, $\widehat{r}_{2}$, $\widehat{r}_{3}$ are maximum values of the ranks $r_1$, $r_2$, $r_3$, respectively (the ranks for the tensor-train decomposition are defined in Eqn.~\ref{eq:tensor_train_decomposition}).

The results are shown on the figure~\ref{fig:level_1}. The best accuracy versus compression ratio is achieved by the method based on CP-decomposition (\cite{lebedev2014speeding}) across all four models. The second best method across all the considered models is the Spatial SVD decomposition~(\cite{jaderberg2014speeding}). We conjecture that good performance of both methods is due to the highly efficient final architecture that is based on horizontal and vertical filters that require few MAC units. In the case of CP-decomposition, the resulting CNN architecture is based on depth-wise separable convolutions, which results in even more savings in computational complexity.

The ranking of the other three methods depends on the model. Thus, choosing the optimal method requires empirical validation. The results show that using higher-level decomposition such as Tucker or tensor-train does not necessarily lead to better performance compared to approaches based on matricization such as weight SVD or spatial SVD.




\subsection{Level 2 compression}

In this section, we present the results of the ablation study for Level 2 methods from~\cite{zhang2016accelerating}, and compare it to channel pruning suggested by~\cite{he2017channel} for Resnet18, and VGG models pre-trained on ImageNet. For data-aware reconstruction, we use 5000 images. For each image, ten $k\times k$ feature map patches at random locations were sampled.

\begin{figure}[t]
\centering
\begin{small}
\begin{tabular}{cc}
\hspace{-0.3cm}\includegraphics[width=7.5cm]{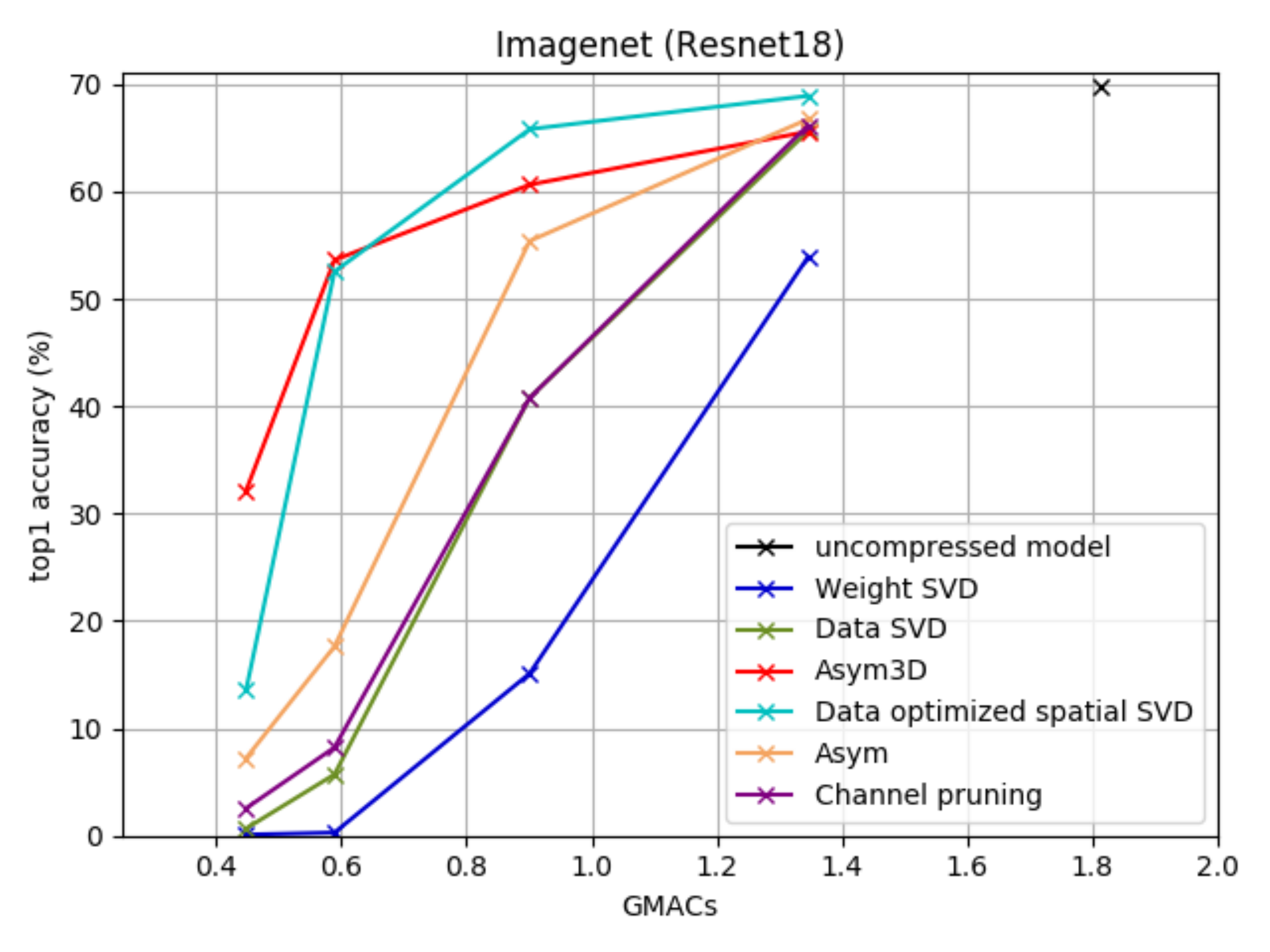} &
\vspace{0.0cm}\includegraphics[width=7.5cm]{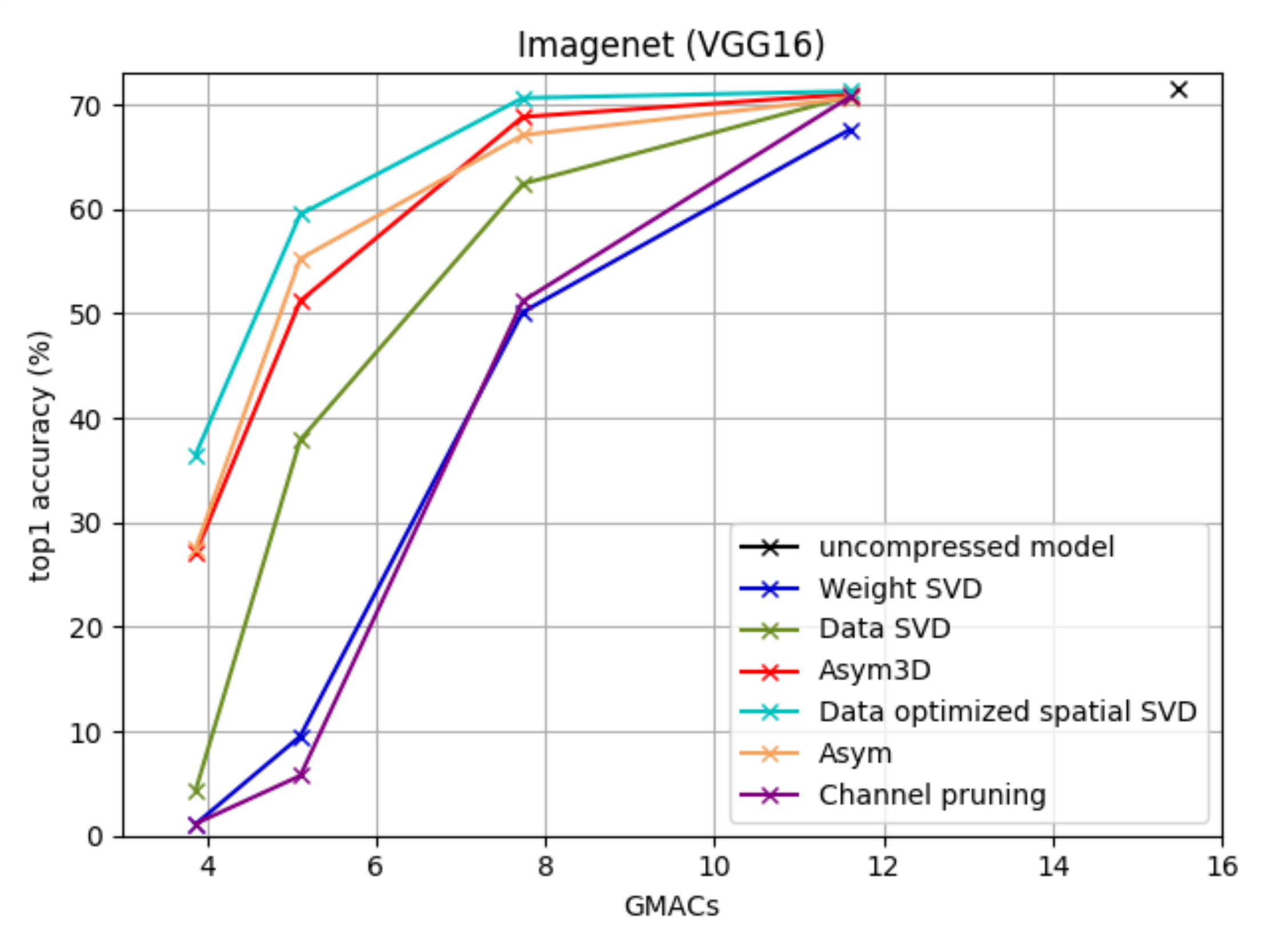} \\
\end{tabular}
\end{small}
\caption{Level 2 compression. Comparison of different data-optimized SVD approaches for Resnet18 trained on ImageNet. The best performance for Resnet18 and VGG16 is achieved using data-optimized spatial SVD.}
\label{fig:level_2_resnet_vgg}
\end{figure}

For the Resnet18 model, five methods were evaluated, including data SVD, asymmetric data SVD, channel pruning, Asym3D, and data-optimized spatial SVD. The best performance for lower compression ratios such as 1.33x and 2x compression is achieved with data-optimized spatial SVD, whereas for higher compression ratios including 3x and 4x compression, better accuracy is achieved using Asym3D (see figure~\ref{fig:level_2_resnet_vgg} on the left). 

The data-optimized spatial SVD method can be seen as three improvements on top of the most basic Level 1 weight SVD compression. The first step is using data for per-layer optimization of the compressed model (Eqn.~\ref{eq:data_svd}), the second is asymmetric formulation (Eqn.~\ref{eq:data_svd_asym}), and finally, some improvement is obtained by using efficient spatial SVD architecture (Eqn.~\ref{eq:gsvd_spatial_opt}). In order to compare improvements due to each step, we performed the following ablation study. As results in figure~\ref{fig:level_2_resnet_vgg} suggest, all three steps are equally important for the compressed model performance.

 The accuracy of channel pruning is mostly on par or comparable with the data SVD method (figure~\ref{fig:level_2_resnet_vgg}) as both methods use data-aware reconstruction based on the same amount of data without leveraging the asymmetric formulation in Eqn.~\ref{eq:opt_gsvd}.

\begin{figure}[H]
\centering
\begin{small}
\includegraphics[width=7.5cm]{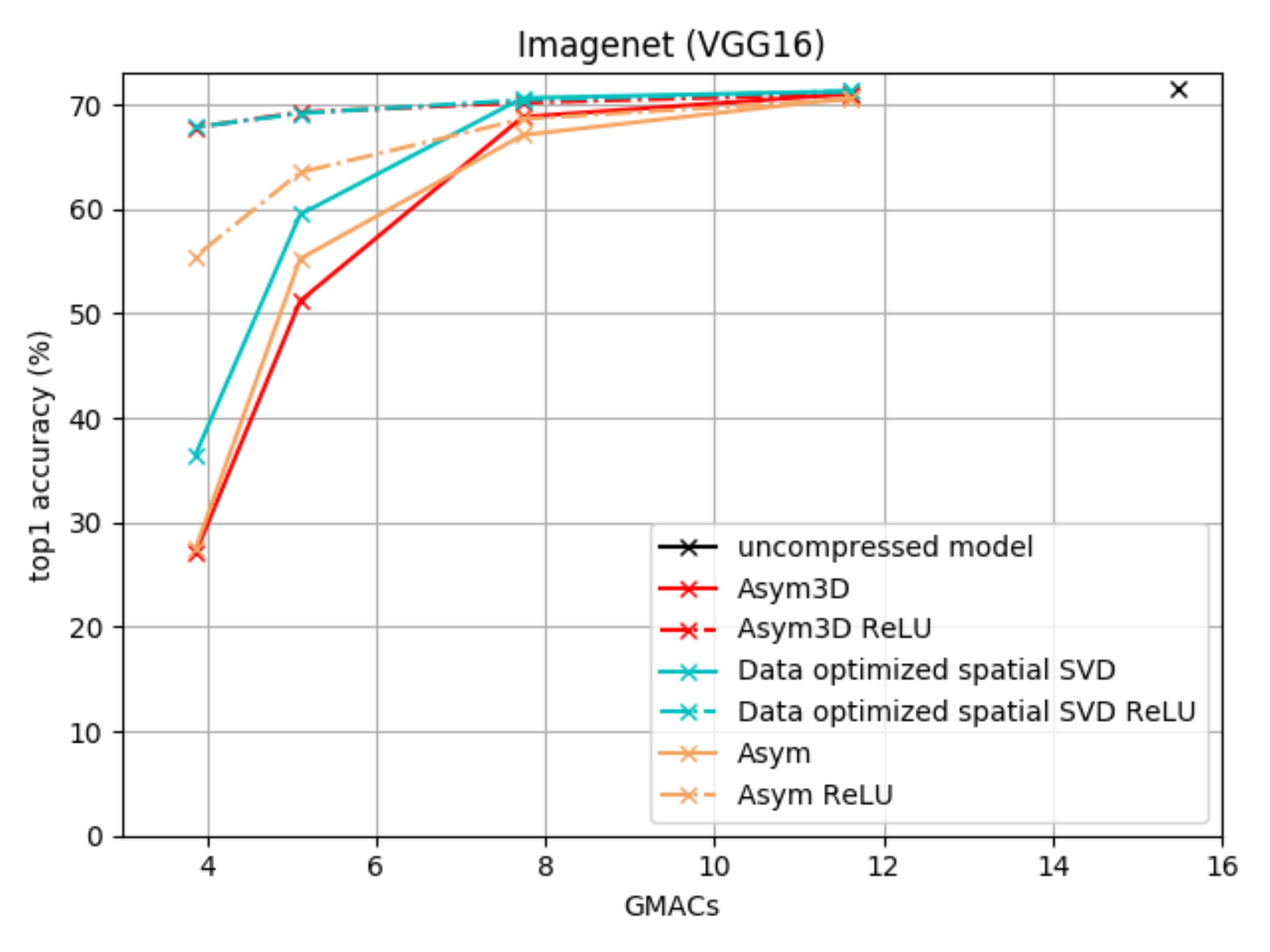}
\end{small}
\caption{Level 2 compression of VGG16 with ReLU activation function included in the formulation.}
\label{fig:level_2_vgg_relu}
\end{figure}

As the VGG16 model has many convolutional layers followed by ReLU non-linearities without batch normalization in between, this model allows adding activation function into the data-aware reconstruction for methods such as asymmetric data SVD, Asym3D, and data-optimized Spatial SVD. The results with the activation function included into the formulation are presented in figure~\ref{fig:level_2_vgg_relu}. The most important part of the methods is using the ReLU function in the optimization, which is necessary for the performance of both methods. 

Overall for VGG16 model, Similar to the Resnet18 results, the three improvements on top of level 1 compression, such as using data-aware optimization, the asymmetric formulation, and using efficient Spatial SVD architecture are equally crucial for the accuracy of the compressed model. For this model, channel pruning demonstrates poor performance, which is comparable to level 1 compression using the weight SVD method.

\begin{figure}[t]
\centering
\begin{small}
\includegraphics[width=7.5cm]{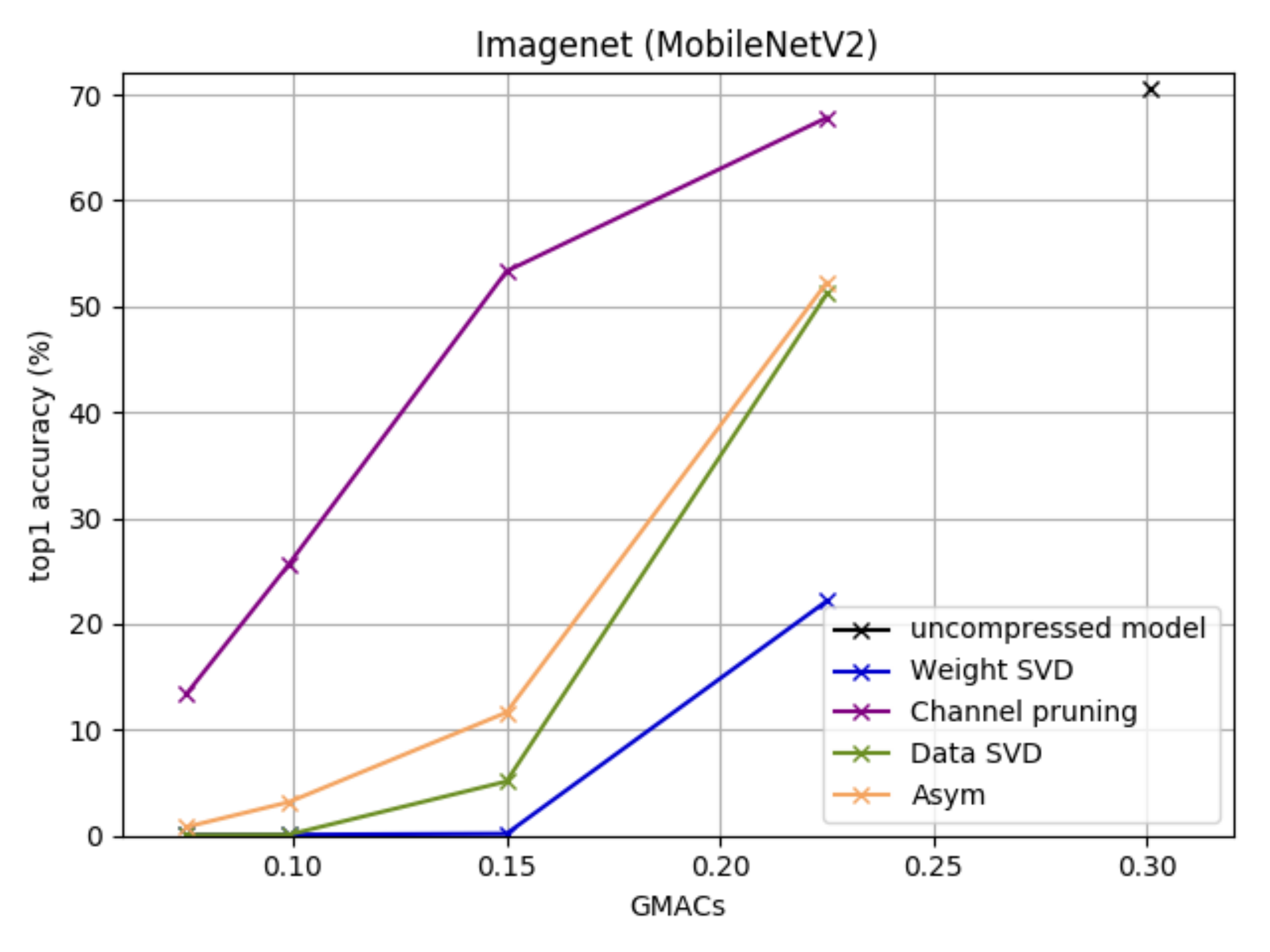}
\end{small}
\caption{Level 2 compression of MobileNet V2. The best performance is obtained using channel pruning, which is applicable to both depth-wise separable and point-wise convolutions. As data-optimized SVD decomposition is only applicable to 1x1 point-wise layers, the performance of this family of methods is lower than channel pruning. However, using data-aware optimization still allows to improve the results of level 1 compression with weight SVD.}
\label{fig:level_2_mobilenetv2}
\end{figure}

The MobileNet V2 architecture is based on depth-wise separable convolutions; therefore, spatial SVD is not possible, and the set of applicable compression methods is restricted to variants of data-optimized SVD, and channel pruning. As the SVD decomposition can only be used for 1x1 convolutional layers and is not applicable for depth-wise separable convolutions, data-optimized SVD methods, including data SVD, asymmetric data SVD, demonstrate poor performance (figure~\ref{fig:level_2_mobilenetv2}) which is still better than data-free weight SVD method. In contrast, channel pruning is applicable for both types of layers, which leads to better accuracy of the compressed model.

\subsection{Level 3 compression}

\subsubsection{Fine-tuned SVD and tensor decompositions} To recover the performance of compressed models, we used the same fine-tuning scheme for different compression methods, the summary for each model is given in the table~\ref{tab:finetuning_schemes}. All the models were fine-tuned using SGD with 0.9 momentum for 20 epochs with learning rate dropped at epochs 10 and 15. Different hyperparameters for each model, including learning rate, batch size, and weight decay value, are given in the table~\ref{tab:finetuning_schemes}.

\begin{table*}
  \centering
  \begin{tabular}{|c | c | c | c | c | c| c|}
    \hline
    Model & learning rate & batch size & weight decay \\ \hline
    Resnet-18 & 0.01 &  256 & $10^{-4}$ \\ \hline
    Resnet-50 & 0.01 & 64 & $10^{-4}$ \\ \hline
    VGG16 & 0.001 & 64 & $10^{-4}$ \\ \hline    InceptionV3 &  0.001 & 64 & $10^{-4}$ \\ \hline
    MobileNetV2 & 0.007 & 256 & $2.0 \times 10^{-4}$ \\ \hline
\end{tabular}
\caption{Fine-tuning schemes for different models trained on ImageNet dataset. All the models were fine-tuned using SGD with 0.9 momentum for 20 epochs with learning rate dropped at epochs 10 and 15.}
\label{tab:finetuning_schemes}
\end{table*}

\begin{figure}[ht]
\centering
\begin{small}
\begin{tabular}{cc}
\hspace{-0.25cm}\includegraphics[width=7.5cm]{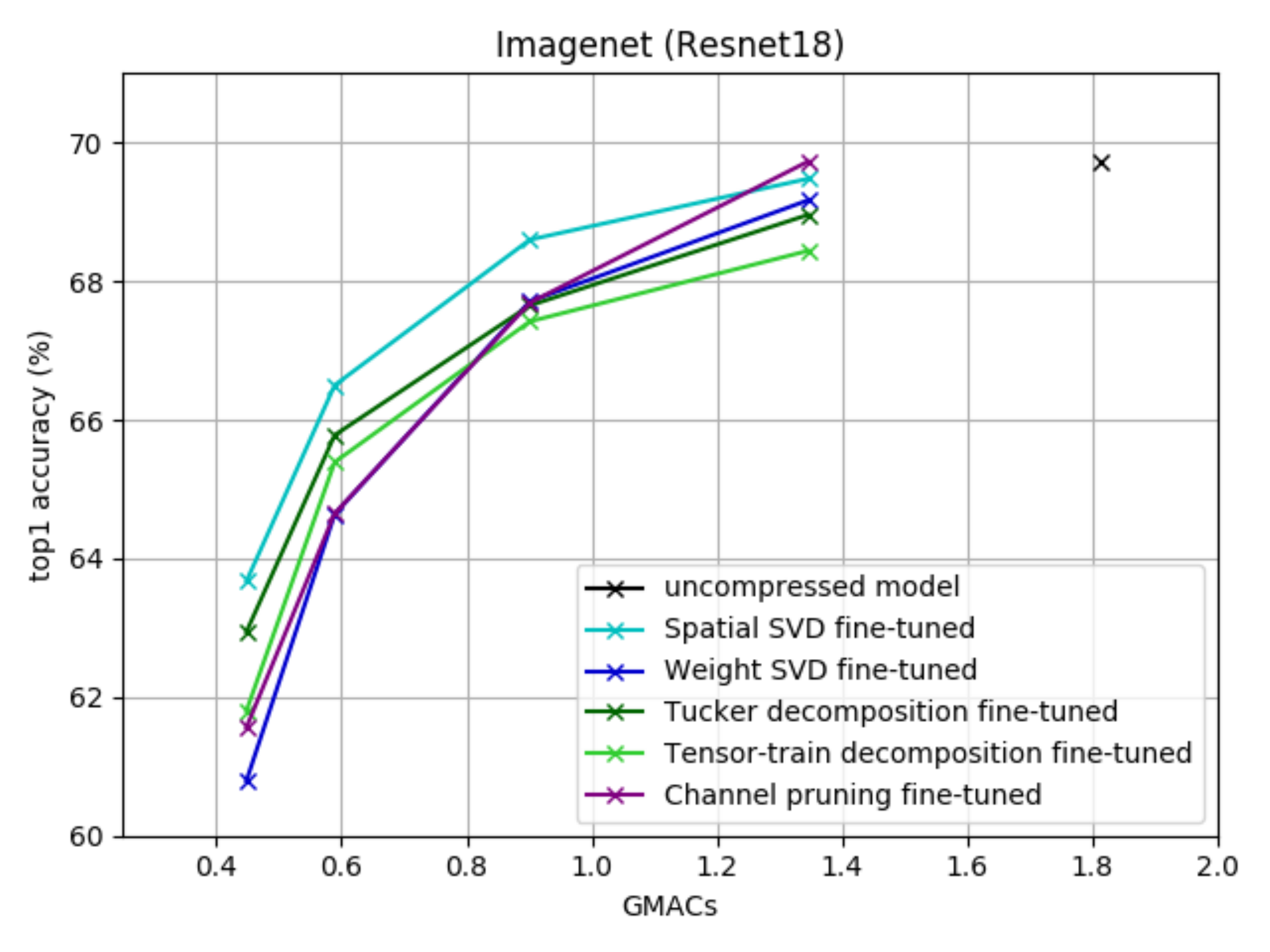} &
\hspace{-0.3cm}\includegraphics[width=7.5cm]{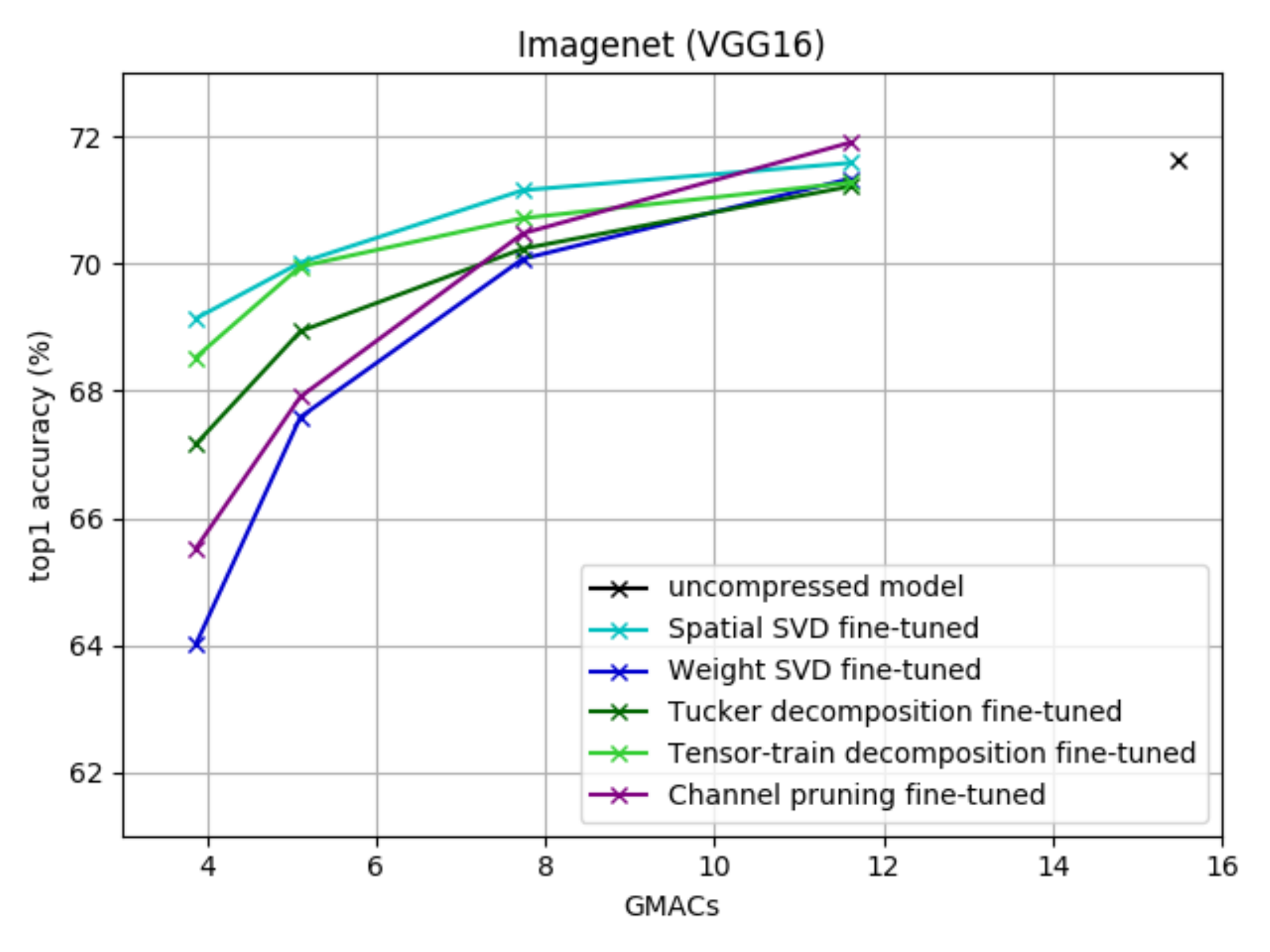}\\
\hspace{-0.3cm}\includegraphics[width=7.5cm]{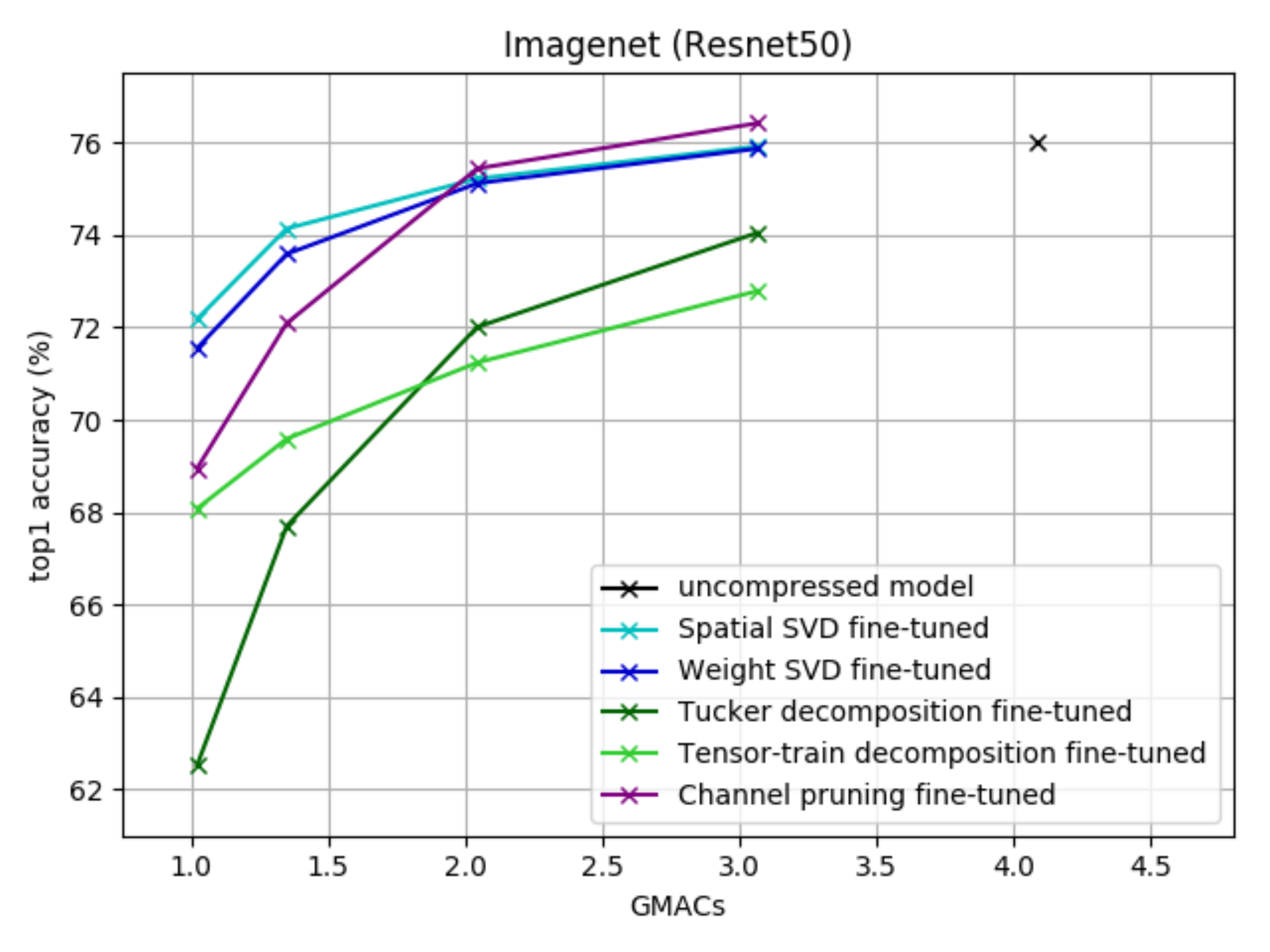} &
\hspace{-0.4cm}\includegraphics[width=7.5cm]{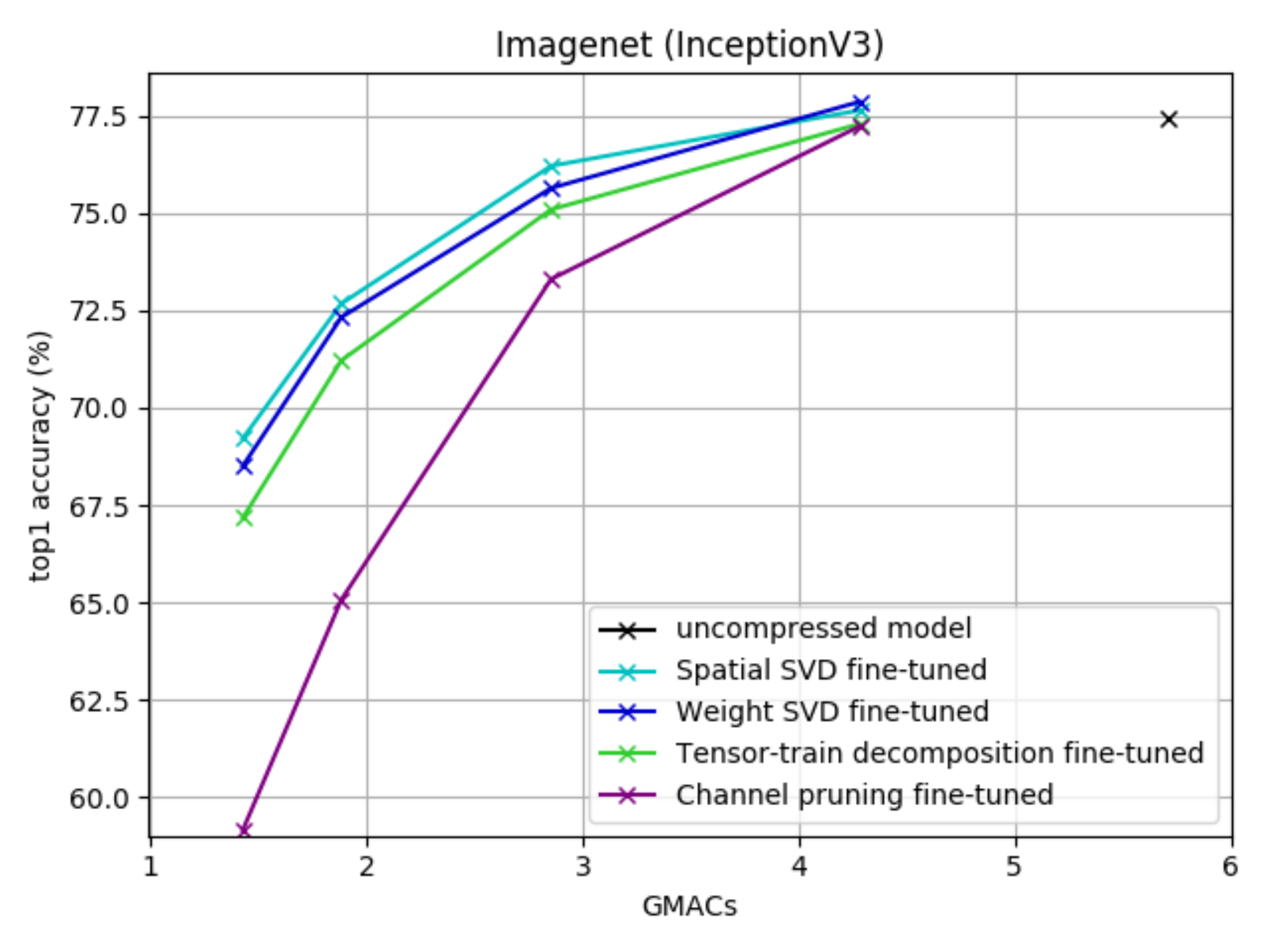}
\end{tabular}
\end{small}
\caption{Level 3 compression. Comparison of different SVD and tensor decomposition methods for Resnet18 trained on ImageNet. The best accuracy is achieved with spatial SVD method across all four models. Ranking of the other methods is different for each specific model.}
\label{fig:level_3}
\end{figure}

\begin{figure}[t]
\centering
\begin{small}
\includegraphics[width=7.5cm]{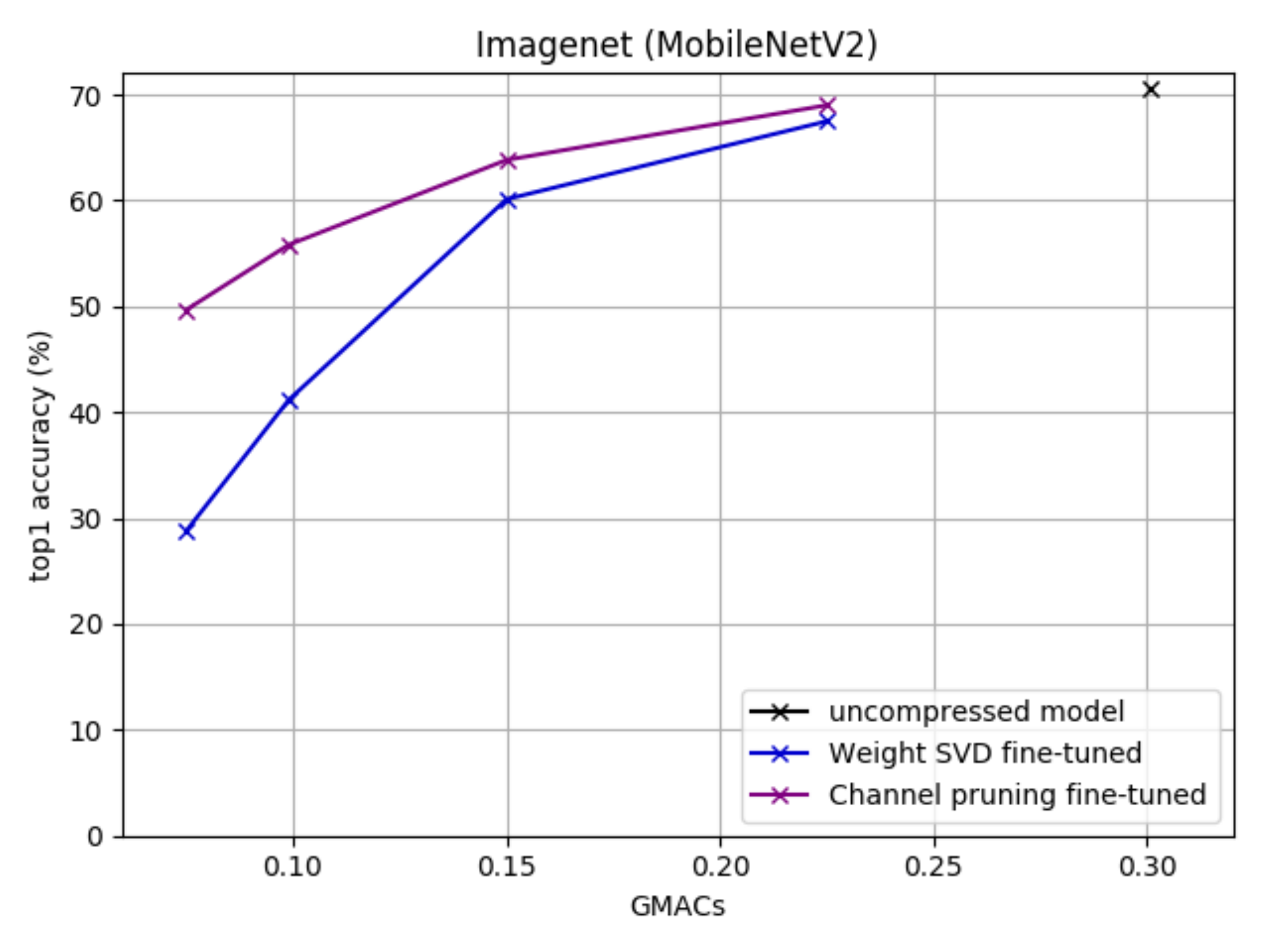}
\end{small}
\caption{Level 3 compression of MobileNetV2. Channel pruning gives better accuracy after fine-tuning compared to weight SVD for all the compression rates.}
\label{fig:level_3_mobilenetv2}
\end{figure}

In figure~\ref{fig:level_3} we show the results for level 3 compression of Resnet18, Resnet50, VGG16, and InceptionV3. The best accuracy for all models is achieved using spatial SVD decomposition. The CP-decomposition shows the best results before fine-tuning. However, the fine-tuning scheme used for all the other methods does not recover the accuracy after compression. We were not able to find any fine-tuning hyperparameters that would allow us to recover the model accuracy. This observation agrees with the results from the original paper by~\cite{lebedev2014speeding}. The results for level 3 compression of MobileNetV2 are given in figure~\ref{fig:level_3_mobilenetv2}. There are only two methods applicable, and channel pruning outperforms fine-tuned weight SVD across all the compression ratios.

Comparing the results for level 1 and level 3 (see figures \ref{fig:level_1} and \ref{fig:level_3}, respectively) suggests that the ranking of compression methods depends on the level, i.e., the best level 1 compression method does not necessarily correspond to the best level 3 compression method.

\subsubsection{Fine-tuning data-optimized SVD}

The following experiment was performed to estimate the potential benefit of combining data-aware optimization with full data fine-tuning for SVD methods. We compressed the Resnet18 network using the level 1 spatial SVD method and level 2 data-optimized spatial SVD. The two methods used the same SVD rank values provided by the greedy method based on singular values so that the resulting network architectures are identical. We fine-tuned both models using the same fine-tuning scheme used for level 3 compression.

\begin{figure}[h]
\centering
\begin{small}
\includegraphics[width=8.0cm]{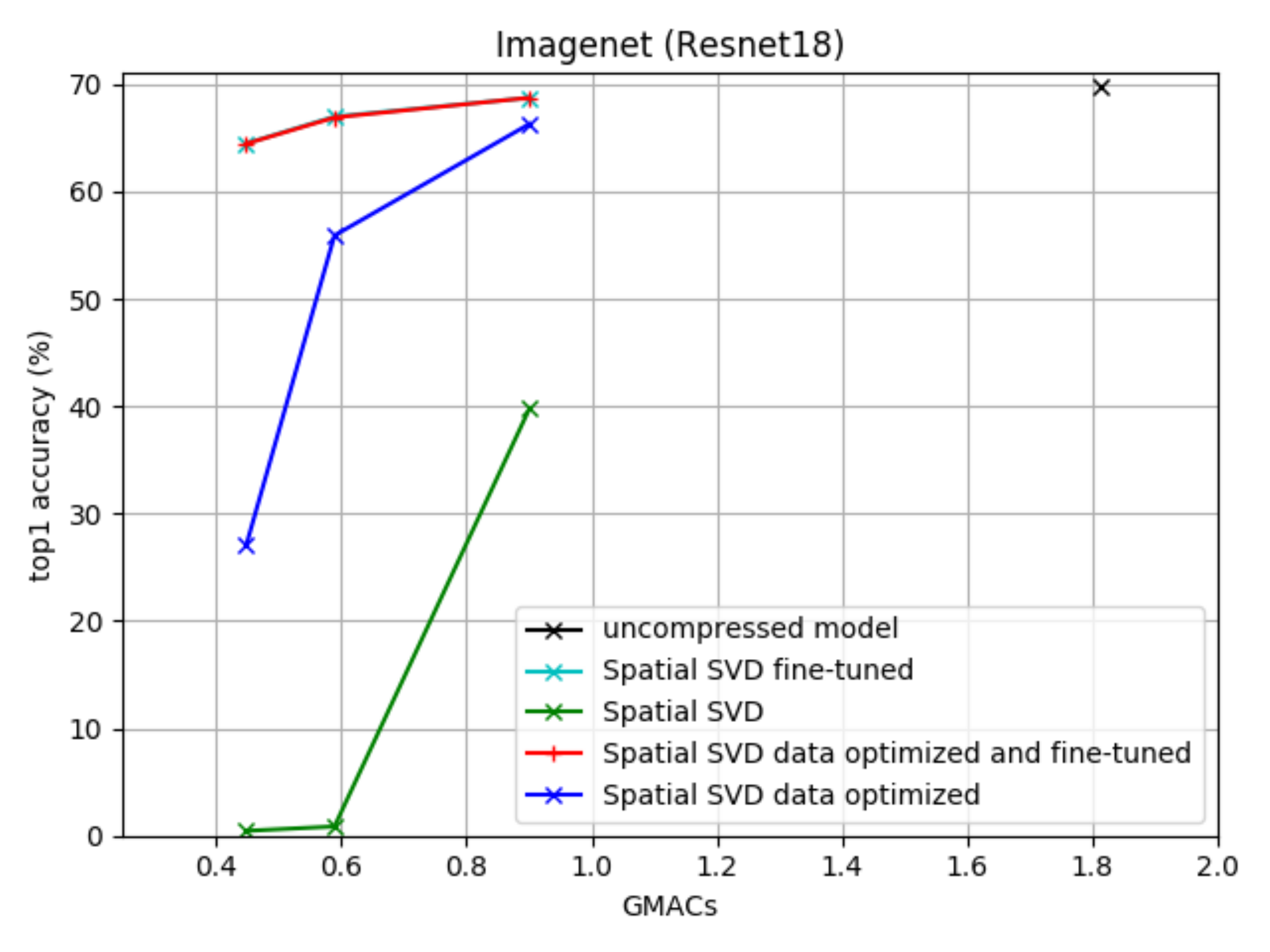}
\end{small}
\caption{Results of performing fine-tuning for a data-optimized compression method. The data-optimized method advantage vanishes after fine-tuning as level 1 spatial SVD method and level 2 spatial method give similar accuracy if fine-tuning is applied after compression. The cyan curve (data-free spatial SVD after fine-tuning) is not visible as it coincides with the red curve (data-optimized spatial SVD after fine-tuning).}
\label{fig:level_2_3}
\end{figure}

The results are shown in figure~\ref{fig:level_2_3}. Despite the substantial difference in accuracy between level 1 and level 2 methods, the difference becomes negligible after the networks are fine-tuned. Therefore, we conclude that there is no benefit in using data-optimized compression if the network is fine-tuned after compression.

\subsubsection{Probabilistic compression}
Contrary to previously discussed methods, methods based on probabilistic compression usually train the network from scratch using a special regularization instead of starting from a pre-trained model\footnote{Probabilistic compression can also be used in combination with a pre-trained model. However, in most cases, this results in lower performance than starting from a randomly initialized model, especially when targeting a high compression rate.}. Since the compression is indirectly enforced using a regularization term and it is not possible to target a specific compression rate directly. However, by line search over regularization strength $\lambda$ we can achieve comparable compression targets than in the other experiments.

Similar to the original model, we train models with probabilistic compression using SGD with a learning rate of 0.1, momentum of 0.9, and a weight decay of $10^{-4}$ for 120 epochs. We drop the learning rate by a factor of 0.1 at epoch 30, 60, and 90. 
The regularization strength $\lambda$ depends on the method applied; for the variational information bottleneck (VIBNet) we used values between $10^{-6}$ and $5\cdot10^{-6}$ to achieve a compression of approximately 1.3x to 3x. For $L_0$ based regularization we used $3\cdot10^{-9}$ to $10^{-8}$ resulting in similar compression rates. 
In case the architecture has residual connections, we add a gate $z$ to the input of the first convolution of each residual block. Thus we can prune the input and output channels of each convolution to achieve an optimal compression rate. Note, in a chain-like CNN pruning the input of a convolution is done implicitly since it depends only on the output of the previous convolution.

\begin{figure}[H]
\centering
\begin{small}
\includegraphics[width=7.5cm]{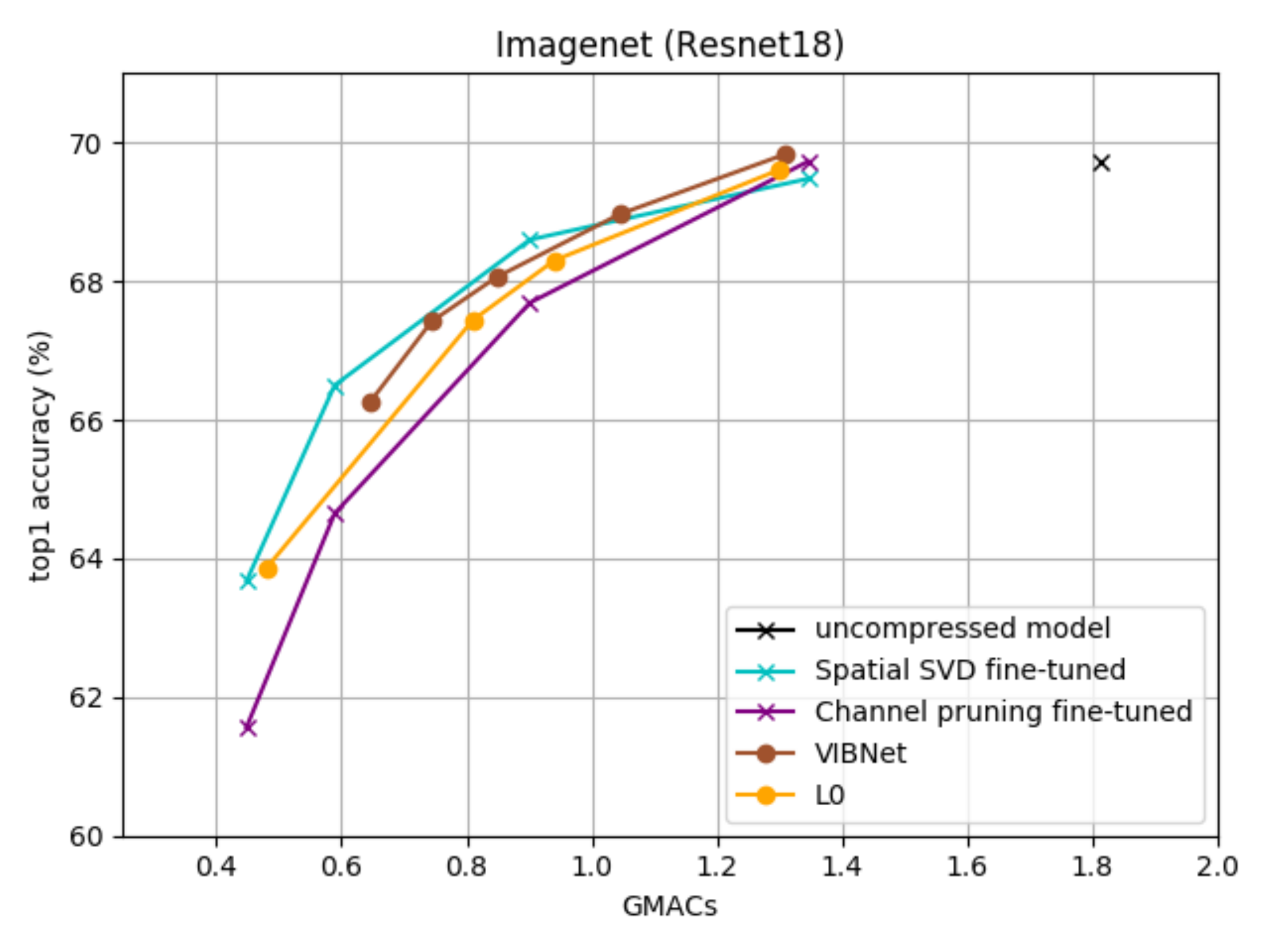}
\end{small}
\caption{Bayesian compression method versus spatial SVD.}
\label{fig:probabilistic}
\end{figure}

The results for probabilistic compression of Resnet18 are shown in figure \ref{fig:probabilistic}. We observe that VIBNet consistently outperforms $L_0$ by a small margin. Compared to the previous best level 3 decomposition method, fine-tuned spatial SVD, VIBNets have a slight edge for lower compression rates but perform worse for very high compression rate. The latter might be due to the fact that spatially decomposing the convolutional filter can lead to a more efficient architecture than only pruning channels. Both VIBNets and $L_0$ consistently outperform fine-tuned channel pruning, which can lead to the same architectures.



\subsubsection{Combining probabilistic compression and channel pruning with SVD compression}

\begin{figure}[H]
\centering
\begin{small}
\includegraphics[width=7.5cm]{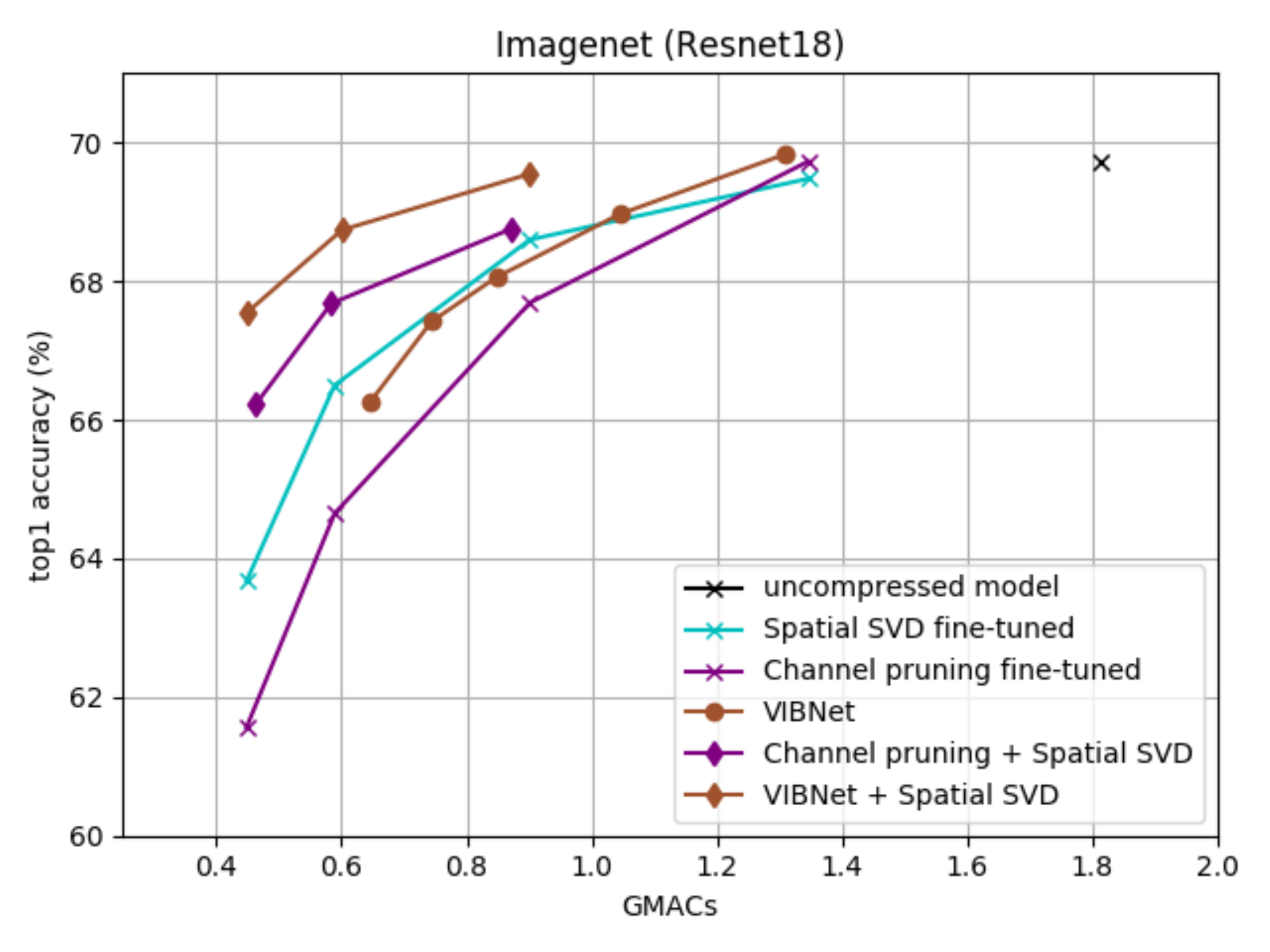}
\end{small}
\caption{Combinations of level 3 compression methods. The best model accuracy is achieved using combining VIBNets trained from scratch with spatial SVD compression. Another practically useful combination is channel pruning applied for the model compressed with spatial SVD. Both combinations allow to improve performance of level 3 compression.}
\label{fig:level3_combinations}
\end{figure}

We found that different level 3 compression approaches are complementary. In fact, spatial SVD can be combined with channel pruning or probabilistic compression, which yields better model accuracy compared to compression using a single full-data method.

The results for the combinations of the methods are given on the figure~\ref{fig:level3_combinations}. In the first case, spatial SVD was applied after probabilistic compression with the VIBNet approach. The VIBNet compressed model was trained from scratch, then spatial SVD was applied for the resulting model, and finally, the compressed model was fine-tuned using the scheme from table~\ref{tab:finetuning_schemes}. 

In the second case, channel pruning was applied after the spatial SVD. After each compression step, we fine-tuned the network with the scheme from table~\ref{tab:finetuning_schemes}. The combination of the VIBNet approach and the spatial SVD achieves the best results, and allows to significantly improve the spatial SVD method.

\subsubsection{Compression versus training from scratch}
One of the important questions related to compression is whether a compressed model gives better performance than training the same architecture from scratch. In order to answer this question, we performed the following experiment. We compressed Resnet18 and VGG16 pre-trained on ImageNet using spatial SVD and channel pruning and then compared the accuracy of the fine-tuned models to the models trained from scratch. The architecture for the models trained from scratch was identical to the architecture obtained by applying the compression techniques. 

The level 3 fine-tuning schemes (table~\ref{tab:finetuning_schemes}) were used for fine-tuning of the compressed models. Whereas for training from scratch, for Resnet18 we use 90 epochs with similar parameters including the starting learning rate 0.1 with dropping it at epochs 30, 60, 90, and for VGG16 62 epochs were used with a learning rate 0.01 dropped at epochs 30, and 60. Using these training parameters for training uncompressed models from scratch gives accuracy equal to the accuracy of the corresponding pre-trained models, which were used for compression. 

\begin{figure}[ht]
\centering
\begin{small}
\begin{tabular}{cc}
\hspace{-0.4cm}\includegraphics[width=7.5cm]{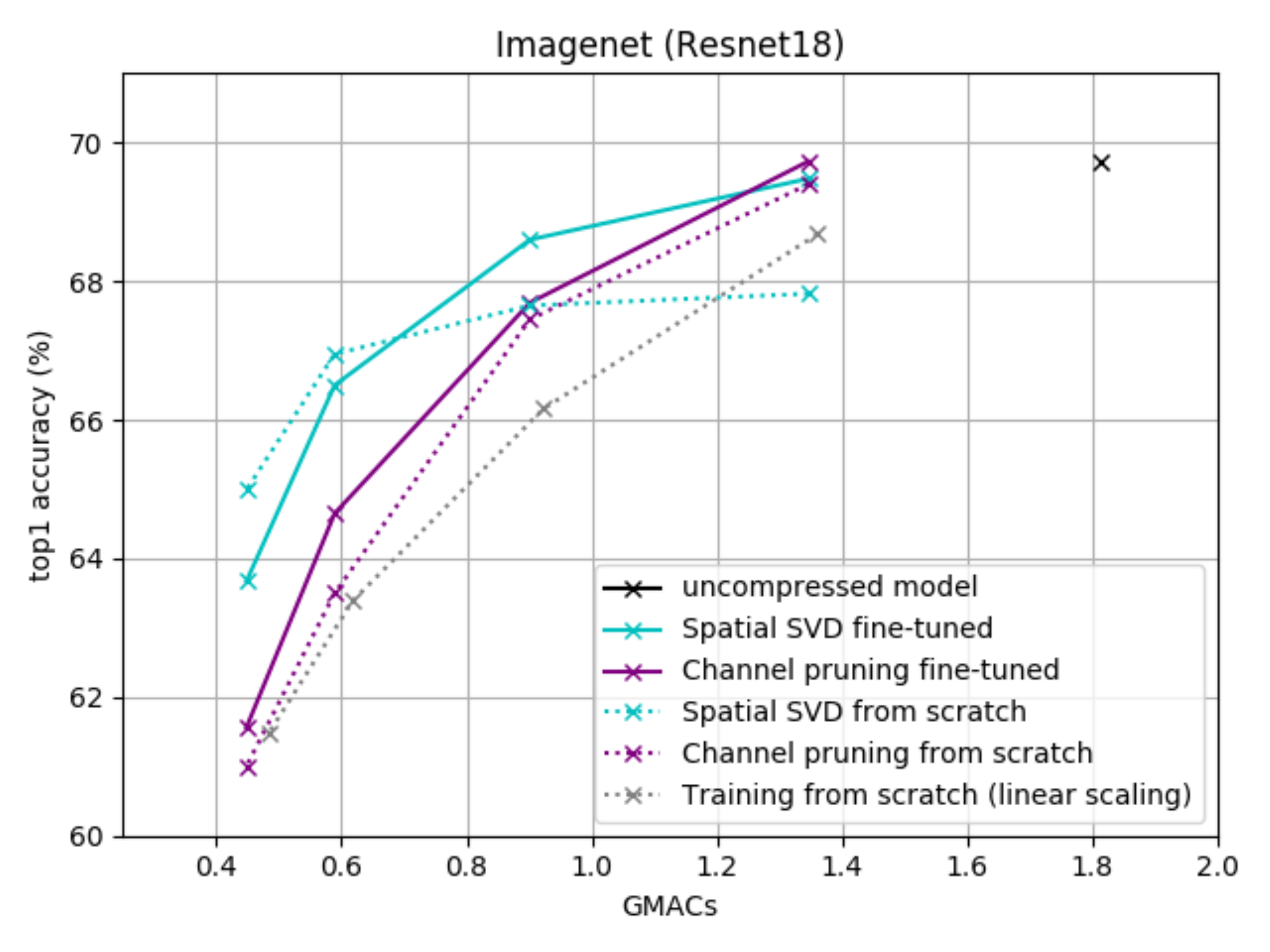} &
\hspace{-0.0cm}\includegraphics[width=7.5cm]{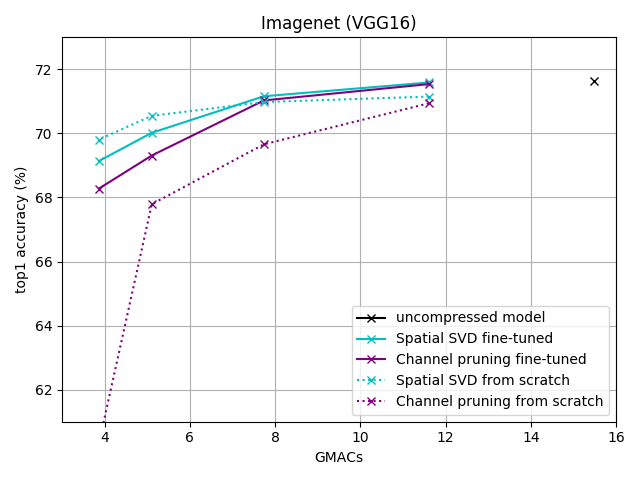}
\end{tabular}
\end{small}
\caption{Full data compression compared to training from scratch for Resnet18, and VGG16 compressed with spatial SVD, and channel pruning. For spatial SVD, training from scratch achieves better accuracy for higher compression rates, and full data compression is more beneficial for moderate compression. For channel pruning a larger model always gives better results than training from scratch.}
\label{fig:experiments_scratch}
\end{figure}

The results are shown in figure~\ref{fig:experiments_scratch}. For channel pruning, using compression always gives better results compared to training from scratch. For spatial SVD using compression outperforms training from scratch for lower compression rates, but training from scratch gives better performance for more aggressive compression. We conjecture that more aggressive compression effectively leaves little information in the pre-trained model. In such cases, training from scratch with random initialization is often better. Our results for lower compression rates agree with the lottery ticket hypothesis~\cite{lottery}, which claims that better accuracy can be achieved by training and pruning a larger model than training a smaller model directly.

\subsection{Compression ratio selection}

One of the important aspects of compression methods is the per layer compression ratio selection. As layers of a network have different sensitivity to compression, different choice of compression ratios can improve or deteriorate the accuracy of the compressed model. The problem of the compression ratio selection can be regarded as a discrete optimization problem. Specifying the full model compression ratio beforehand results in a constraint imposed on the solution. 

\begin{figure}[t]
\centering
\begin{small}
\includegraphics[width=8.0cm]{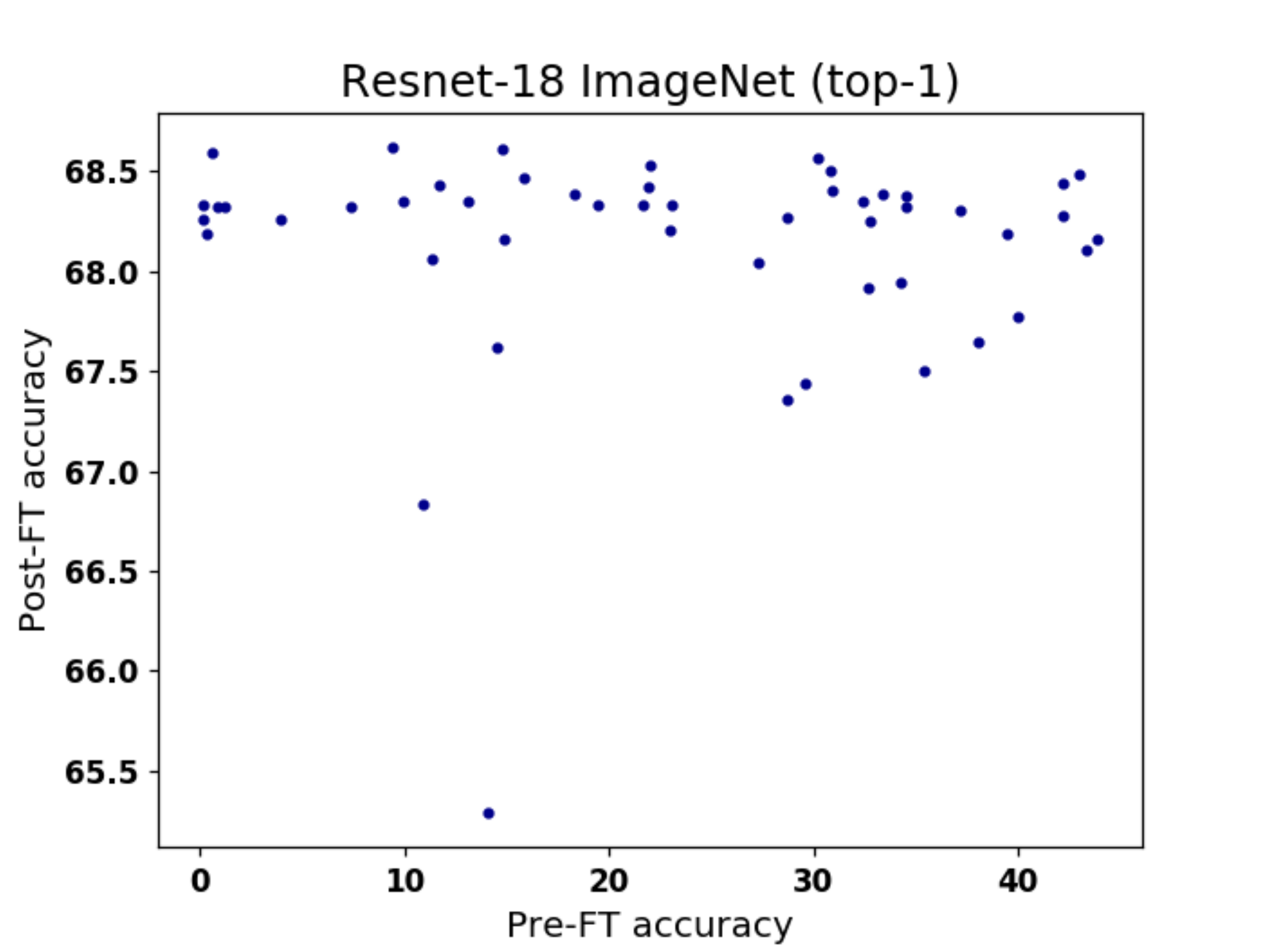}
\end{small}
\caption{Spatial SVD, pre-finetuning accuracy versus post-finetuning accuracy for different sets of SVD ranks. The plot is based on the Resnet18 network compressed using spatial SVD with a 2x compression ratio. We fine-tuned 50 different compressed models with different values of SVD ranks to check whether the pre-finetuning accuracy for each model is a good proxy for its post-finetuning accuracy. All the 50 models have equal MAC count. The results of the experiment suggest that there is no correlation between the two accuracies so that it is not possible to use pre-finetuning accuracy to optimize per-layer compression ratios.}
\label{fig:rank_selection}
\end{figure}

The choice of the objective function for the optimization corresponds to several different practical use cases of model compression. Besides the obvious choice of maximizing the accuracy of the compressed model, compression ratio selection can be used to minimize the inference time of the compressed model on specific hardware leading to hardware-optimized compression. In addition to inference time, the objective function can be based on the memory footprint of the model at inference time as well as use any combination of the quantities mentioned above.

Practical usage of the compression ratio optimization faces a challenge related to the need for time-consuming model fine-tuning to recover the compressed model accuracy. Using model accuracy after fine-tuning as an objective function for optimization is prohibitively expensive in this case. One way to alleviate this problem used in the literature (e.g., \cite{he2018amc}) is the following: a model is compressed using a set of compression ratios and evaluated on the validation set without fine-tuning. Then this accuracy value is being used in the optimization as a proxy of the accuracy of the compressed model after fine-tuning. In this case, it is assumed that a better compressed model accuracy before fine-tuning leads to a better compressed model accuracy after fine-tuning.

To quantitatively validate this assumption, we performed the following experiment. First, we compressed the Resnet18 model with a 2x compression ratio; the compression ratios per layer were selected using the greedy method based on singular values. Second, we randomly perturbed the compression ratios in a way such that the full model complexity is preserved under the perturbations. This way, we obtained 50 different compressed Resnet-18 models of the same computational complexity. To verify whether the model accuracy before fine-tuning is a suitable proxy for the model accuracy after fine-tuning, we fine-tuned all the models using the same fine-tuning scheme, which was used for level 3 compression (see table~\ref{tab:finetuning_schemes}). The figure~\ref{fig:rank_selection} shows the results as a scatter plot with the horizontal axis corresponding to the model accuracy before fine-tuning and vertical axis corresponding to the accuracy after fine-tuning. As the results suggest, there is no correlation between the two accuracy values. This does not agree with the assumption made above and leaves the problem of practical compression ratio optimization for architecture search methods wide open.  



%% file: sections/conclusion.tex
In this paper, we performed an extensive experimental evaluation of different neural network compression techniques. We considered several methods, including methods based on SVD, tensor factorization, channel pruning, and probabilistic compression methods.

We introduced a methodology for the comparison of different compression techniques based on levels of compression solutions. Level 1 corresponds to data-free compression with no fine-tuning or optimization used to improve the compressed model. Level 2 corresponds to data-optimized compression based on a limited number of training data batches used to improve the predictions by performing layer-wise optimization of the parameters of the compressed model. No back-propagation is used at this level. Level 3 corresponds to fine-tuning the compressed model on the full training set using back-propagation. We hope these levels help distinguish between different types of compression methods more clearly, as the vocabulary is adopted.

Experimental evaluation of the considered methods shows that the performance ranking of the considered methods depends on the level chosen for experiments. At level 1, CP-decomposition shows the best accuracy for most of the models. The ranking of the other methods depends on the model. 

The best results for Level 2 compression are achieved using per-layer optimization based on the combination of asymmetric formulation (\cite{zhang2016accelerating}); however, our experiments show that applying the GSVD method for optimizing the second factor of spatial SVD decomposition yields better results than the original Asym3D double decomposition approach from the same paper. 

For level 3 compression, the best performance is given by VIBNet and L0 methods for moderate compression, and by the spatial SVD for higher compression ratios. In additional experiments, we show that SVD compression is complementary to channel pruning and probabilistic pruning approaches so that using the combination of VIBnet and spatial SVD gives the best performance overall of any of the considered compression techniques.

In further experiments, we demonstrate that level 3 compression of a larger network achieves better performance compared to training a smaller network from scratch, both for SVD-based compression, and pruning methods. These results are in agreement with the lottery ticket hypothesis (\cite{lottery}) and indicate that compression methods should be applied after training, and are not just a way of doing neural architecture search.